\documentclass{bmvc2k}
\usepackage{times}
\usepackage{graphicx}
\usepackage{wrapfig,lipsum,booktabs}
\usepackage{xcolor}
\usepackage{colortbl}
\usepackage{textcomp}
\usepackage{gensymb}
\usepackage{array,bm,multirow}
\usepackage{enumitem}

\title{Pose from Shape: Deep Pose Estimation \\ for Arbitrary 3D Objects}
\addauthor{Yang Xiao}{https://youngxiao13.github.io}{1}
\addauthor{Xuchong Qiu}{https://imagine-lab.enpc.fr/staff-members/xuchong-qiu/}{1}
\addauthor{Pierre-Alain Langlois}{http://imagine.enpc.fr/~langloip/}{1}
\addauthor{Mathieu Aubry}{http://imagine.enpc.fr/~aubrym/}{1}
\addauthor{Renaud Marlet}{http://imagine.enpc.fr/~marletr/}{1}

\addinstitution{
	LIGM (UMR 8049)\\
	Ecole des Ponts ParisTech, UPE\\
	Champs-sur-Marne, France
}

\runninghead{Y.XIAO ET AL.}{Deep Pose Estimation for Arbitrary 3D Objects}

\def\etal{\emph{et al}\bmvaOneDot}

\usepackage{amsmath}

\def\clap#1{\hbox to 0pt{\hss#1\hss}}
\newcommand{\azi}{{\mathsf{azi}}}
\newcommand{\ele}{{\mathsf{ele}}}
\newcommand{\inp}{{\mathsf{inp}}}
\newcommand\Euler{\mathcal{E}}
\newcommand\Loss{\mathcal{L}}
\newcommand\class{{\mathsf{cla}}}
\newcommand\reg{{\mathsf{reg}}}
\newcommand\gt{{\mathsf{gt}}}
\newcommand\pred{{\mathsf{pred}}}
\newcommand\shat[1]{\smash{\hat{#1}}}
\newcommand\Accpisix{\mathit{Acc}_{\smash{\frac{\pi}{6}}}}
\newcommand\Accpisixb{\bm{Acc}_{\smash{\frac{\pi}{6}}}}
\newcommand\MedErr{\textit{MedErr}}
\newcommand\ADD{ADD-0.1d}
\newcommand\ADDS{ADD-S-0.1d}

\newcolumntype{L}[1]{>{\raggedright\let\newline\\\arraybackslash\hspace{0pt}}m{#1}}
\newcolumntype{C}[1]{>{\centering\let\newline\\\arraybackslash\hspace{0pt}}m{#1}}
\newcolumntype{R}[1]{>{\raggedleft\let\newline\\\arraybackslash\hspace{0pt}}m{#1}}

\definecolor{Gray}{gray}{0.85}
\definecolor{HL}{rgb}{0.95,1,0.95}
\newcolumntype{a}{>{\columncolor{gray}}c}
\newcolumntype{b}{>{\columncolor{white}}c}

\begin{document}
	
\maketitle
	
\begin{abstract}
Most deep pose estimation methods need to be trained for specific object instances or categories.
In this work we propose a completely generic deep pose estimation approach, which does not require the network to have been trained on relevant categories, nor objects in a category to have a canonical pose.
We believe this is a crucial step to design robotic systems that can interact with new objects ``in the wild'' not belonging to a predefined category. % nor to have a canonical viewpoint. 
Our main insight is to dynamically condition pose estimation with a representation of the 3D shape of the target object. More precisely, we train a Convolutional Neural Network that takes as input both a test image and a 3D model, and outputs the relative 3D pose of the object in the input image with respect to the 3D model. %We represent the 3D models either as a point cloud or set of rendered views. 
We demonstrate that our method boosts performances for supervised category pose estimation on standard benchmarks, namely Pascal3D+, ObjectNet3D and Pix3D, on which we provide results superior to the state of the art. More importantly, we show that our network trained on everyday man-made objects from ShapeNet generalizes without any additional training to completely new types of 3D objects by providing results on the LINEMOD dataset as well as on natural entities such as animals from ImageNet. Our code and model is avalaible at \href{http://imagine.enpc.fr/~xiaoy/PoseFromShape/}{http://imagine.enpc.fr/\textasciitilde xiaoy/PoseFromShape/}.
\end{abstract}

%-------------------------------------------------------------------------
\section{Introduction}
\label{sec:intro}

Imagine a robot that needs to interact with a new type of object not belonging to any pre-defined category, such as a newly manufactured object in a workshop. Using existing single-view pose estimation approaches for this new object would require stopping the robot and training a specific network for this object before taking any further action. Here we propose an approach that can directly take as input a 3D model of the new object and estimate the pose of the object in images relatively to this model, without any additional training procedure. We argue that such a capability is necessary for applications such as robotics ``in the wild'', where new objects of unfamiliar categories can occur routinely at any time and have to be manipulated or taken into account for action. It also applies to virtual reality with similar circumstances.

% Neural Network are impressively good at estimating the viewpoints of chairs or other common object categories in an image. In this paper, we design an approach that can also estimate the viewpoint of a unique object without any additional training. 
% Being able to address the orientation of totally arbitrary objects is not only useful, e.g., for virtual reality, it actually 
% In such a scenario, training a network for this specific object before taking an action is unpractical. % without pausing for training update -- assuming a dataset exists for these new categories, which is unlikely in many scenarios.

\begin{figure}
\centering
    \hspace*{-3mm}
	\begin{tabular}{c@{\hspace*{8mm}}c}
    \bmvaHangBox{\includegraphics[height=58mm]{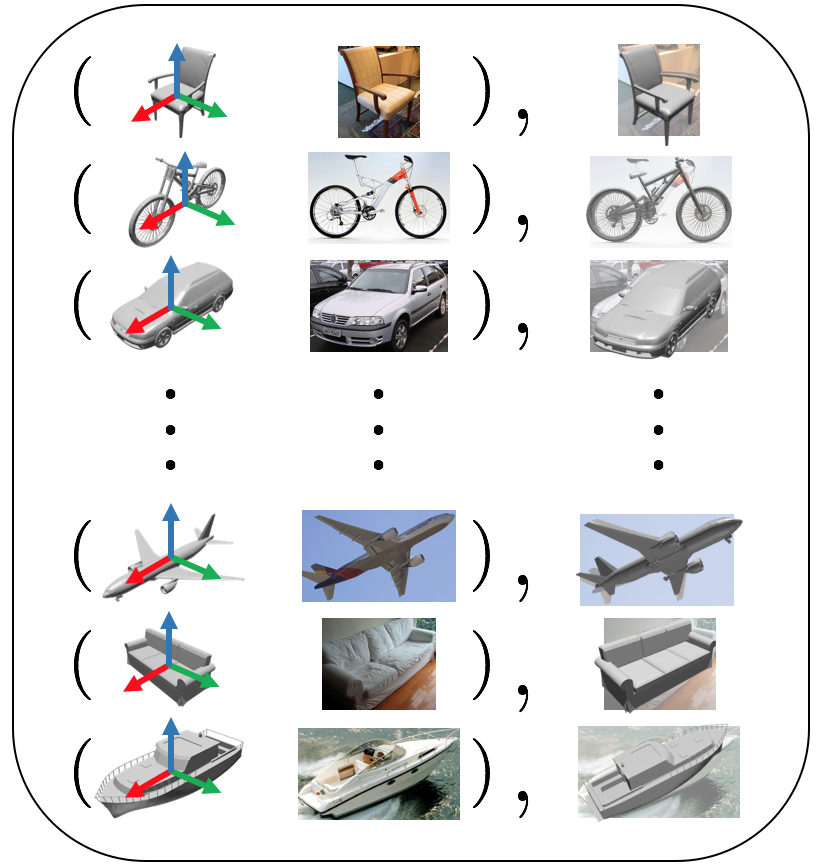}}&
    \bmvaHangBox{\includegraphics[height=58mm]{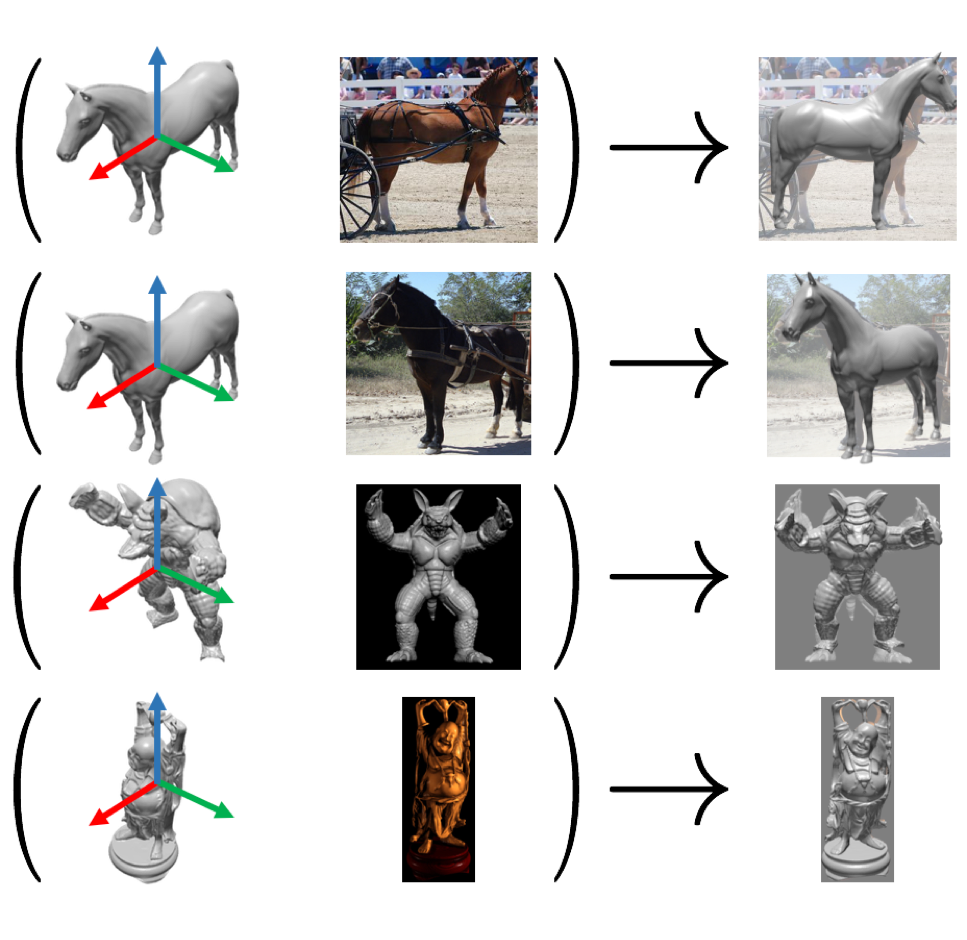}}\\
    (a) Training with shape and pose &(b) Testing on unseen objects
    \end{tabular}
    \vspace{2mm}
    \caption{Illustration of our approach. (a) Training data: 3D model, input image and pose annotation for everyday man-made object;
    % (a) Examples of training samples (3D model, RGB image of object, rendering of the model under the ground truth pose);
    % For each training sample, we show: the 3D shape; the RGB image and rendering of the ground truth model under the ground truth pose. 
    %(b) During testing, we can retrieve the pose of any arbitrary object given the RGB image and its shape.}
    (b) At testing time, pose estimation of any arbitrary object, even an unknown category, given a RGB image and the corresponding 3D shape.}
    \vspace{-3mm}
\label{fig:teaser}
\end{figure}

% While existing methods achieved good results on standard datasets, they lacked the generalization ability towards unseen objects that are not included in the training data. This strongly limits the applicability of deep learning methods in scenarios where the shape is known, but does not belong to any annotated category.
To overcome the fact that deep pose estimation methods were category-specific, i.e., predicted different orientations according to object category, recent works~\cite{3Dpose3Dmodel2018,starmap2018} have proposed to perform category-agnostic pose estimation on rigid objects, producing a single prediction. However, \cite{3Dpose3Dmodel2018} only evaluated on object categories that were included in the training data, while~\cite{starmap2018} required the testing categories to be similar to the training data. On the contrary, we want to stress that our method works on novel objects that can be widely different from those seen at training time.  For example, we can train only on man-made objects, but still be able to estimate the pose of animals such as horses, whereas not a single animal has been seen in the training data (cf.\ Fig.~\ref{fig:teaser} and~\ref{fig:visual_results}). Our method is similar to category-agnostic approaches in that it only produces one pose prediction and does not require additional training to produce predictions on novel categories. However, it is also instance-specific, because it takes as input a 3D model of the object of interest. 

Indeed, our key idea is that viewpoint is better defined for a single object instance given its 3D shape than for whole object categories. Our work can be viewed as leveraging the recent advances in deep 3D model representations~\cite{mvcnn2015,Qi2017PointNetDL,Qi2017PointNetDH} for the problem of pose estimation. We show that using 3D model information also boosts performances on known categories, even when the information is only approximate, as in the Pascal3D+~\cite{pascal3d2014} dataset. %, ObjectNet3D~\cite{object3d2016} and Pix3D~\cite{pix3d2018} datasets

When an exact 3D model of the object is known, as in the LINEMOD~\cite{hinterstoisser2012LINEMOD} dataset, state-of-the-art results are typically obtained by first performing a coarse viewpoint estimation and then applying a pose-refinement approach, typically matching rendered images of the 3D model to the target image. Our method is designed to perform the coarse alignment. 
% More precisely, we focus on the problem of coarse pose estimation in the case where an approximate 3D model of the object of interest is known. Given a single image of an object, we estimate the coarse orientation of the camera with respect to an arbitrary frame of reference on the 3D model. Such an estimation can then be refined by pose-refinement methods, 
Pose-refinement can be performed after applying our method using a classical approach based on ICP or the recent DeepIM~\cite{li2018deepim} method. Note that while DeepIM only performs refinement, it is similar to our work in the sense that it is category agnostic and leverages some knowledge of the 3D model, using a view rendered in the estimated pose, to predict its pose update.

% % \todo{should we mention/insist on single view?}
% In this paper, we demonstrate how to estimate the pose of an arbitrary object by training a deep network on a collection of man-made objects. This is in contrast to most deep learning approaches which have boosted performances in object pose estimation on standard datasets by relying on training networks to predict pose for specific object categories. This dependency can be either explicit, using one network or one branch per category, or implicit, because the reference pose is defined in a different way for each category. This strongly limits the applicability of such approaches in scenario where the test shape is known, but does not belong to any annotated category. In the case where the network predicts a single pose, it also makes its task more complex, since it has to implicitly perform recognition, as the reference viewpoint might be defined differently for different object categories and the orientation.

Our core contributions are as follows:
\begin{itemize}%[topsep=0pt,itemsep=0pt]
	\item To the best of our knowledge, we present the first deep learning approach to category-free viewpoint estimation, which can estimate the pose of any object conditioned only on its 3D model, whether or not it is similar to objects seen at training time.
	\item We can learn with and use ``shapes in the wild'', whose reference frame do not have to be consistent with a canonical orientation, simplifying pose supervision. %\mathieu{, thus requiring less supervision and allowing simultaneous training on heterogenous datasets. -> I don't really understand the "less supervision" -  I guess it referred to the category, but since we require the 3D model, we also require more supervision - and training on heterogenous dataset seems weak to me: ccl, I think the contribution is stronger without the second part of the sentence }
	\item We demonstrate on a large variety of datasets~\cite{pascal3d2014,object3d2016,pix3d2018,hinterstoisser2012LINEMOD} that adding 3D knowledge to pose estimation networks provides performance boosts when applied to objects of known categories, and meaningful performances on previously unseen objects. %our category-free pose estimation, when applied to previously unseen objects,  has similar performances as category-specific methods. we also show
\end{itemize}

%-------------------------------------------------------------------------
\section{Related Work}
\label{sec:relawork}

%Since there is a vast amount of literature on pose estimation, we present representative relative works for object pose estimation where input is object RGB image with or without using its 3D shape.
In this section, we discuss pose estimation of a rigid object from a single RGB image first in the case where the 3D model of the object is known, then when the 3D model is unknown.
\vspace{-3mm}

\paragraph{Pose estimation explicitly using object shape.}
Traditional methods to estimate the pose of a given 3D shape in an image can be roughly divided into feature-matching methods and template-matching methods. 
%The object shape representation has many variants such as 3D mesh, voxel, point cloud or synthetic rendered image using model.
Feature-matching methods try to extract local features from the image, match them to the given object 3D model and then use a variant of PnP algorithm to recover the 6D pose based on estimated 2D-to-3D correspondences. Increasingly robust local feature descriptors~\cite{lowe2004distinctive,tola2010daisy,ViewpointsKeypoints2015,pavlakos20176}  and more effective variants of PnP algorithms~\cite{lepetit2009epnp,zheng2013revisiting,li2012robust,ferraz2014very} have been used in this type of pipeline.  Pixel-level prediction, rather than detected features, has also been proposed~\cite{Brachmann2016UncertaintyDriven6P}. Although performing well on textured objects, these methods usually struggle with poorly-textured objects. To deal with this type of objects, template-matching methods try to match the observed object to a stored template~\cite{li2011deformable,lowe1991fitting,hinterstoisser2012gradient,hinterstoisser2012LINEMOD}. However, they perform badly in the case of partial occlusion or truncation. % \mathieu{REMOVE?: It is also possible to try to directly predict the 3D coordinates of object model vertices for each pixel and infer the pose~\cite{Brachmann2016UncertaintyDriven6P}}\xuchong{as it's compared later, keep it here is better?}.

More recently, deep models have been trained for pose estimation from an image of a known or estimated 3D model. % to predict some keypoint postions. %, either in 2D or directly in 3D. 
%with supervised data to find object local features or to offer better template representations. Some work 
% A first set of works predicts the position of keypoints. 
Most methods estimate the 2D position in the test image of the projections of the object 3D bounding box~\cite{rad2017bb8,tekin2018real,oberweger2018heatmapspose, 3Dpose3Dmodel2018} or object semantic keypoints~\cite{pavlakos20176,MatchingRI2018} to find 2D-to-3D correspondences and then apply a variant of the PnP algorithm, as feature-matching methods. %similarly to the feature-matching methods described above. 
Once a coarse pose has been estimated, deep refinement approaches in the spirit of template-based methods have also been proposed~\cite{manhardt2018modelrefine,li2018deepim}. % Following a coarse pose estimation, given object shape also enable a pose refinement module using ICP or more sophisticated methods. Meanwhile, recently some independent deep pose refinement methods using object shape have also been proposed~\cite{manhardt2018modelrefine,li2018deepim}. 

\vspace{-3mm}
\paragraph{Pose estimation not explicitly using object shape.}
In recent years, with the release of large-scale datasets~\cite{Geiger2012kitti, hinterstoisser2012LINEMOD, pascal3d2014,object3d2016,pix3d2018}, data-driven learning methods (on real and/or synthetic data) have been introduced which do not rely on an explicit knowledge of the 3D models. %On these datasets, it is possible to train deep network to predict object pose without having any explicit knowledge of the 3D model of the objects. %, learning-based approaches using
% only on example images without object shape information became possible and proved its effectiveness. Conventionally, the model is trained on supervised data to directly map the observed object image to object pose representations with reference to its reference pose. 
These can roughly be separated into methods that estimate the pose of any object of a training category and methods that focus on a single object or scene. 
For category-wise pose estimation, a canonical view is required for each category with respect to which the viewpoint is estimated. The prediction can be cast as a regression problem~\cite{Osadchy2007,Penedones2012ImprovingOC,MultiTask2016}, a classification problem~\cite{ViewpointsKeypoints2015,RenderForCNN2015,Elhoseiny2016ACA} or a combination of both~\cite{3DBboxMousavian2017,Guler2017,UnifiedMVMC2018,Mahendran2018AMC}. Besides, Zhou \etal directly regress category-agnostic 3D keypoints and estimate a similarity between image and world coordinate systems~\cite{starmap2018}.
 %Besides, Zhou \etal directly regress category-agnostic 3D keypoints of observed view/canonical view and then resolve for a similarity transformation to recover the observed viewpoint~\cite{starmap2018}.  
%single instance 
Following the same strategy, it is also possible to estimate the pose of a camera with respect to a single 3D model but without actually using the 3D model information. Many recent works have applied this strategy to recover the full 6-DoF pose for object~\cite{RealTimeMP2017, 3DBboxMousavian2017,SSD6D2017,PoseCNN2018,UnifiedMVMC2018} and camera re-localization in the scene~\cite{posenet2015, kendall2017geometricpose}. % All the methods described in this paragraph can be trained with real annotated images, synthetic rendered views of the 3D models, or a mix of both.

%SSD-6D does prediction without object shape, but do refinement with shape. I added it in Pose Estimation not explicitly using object shape
%
% \paragraph{Generalization to unseen objects}
% Paul Wohlhart \etal use depth map as input to output a feature descriptor and do descriptor-matching in database to recover unseen object class and relevant pose~\cite{wohlhart2015learning}. Balntas \etal evaluate their pose guied RGBD faeture learning on unseen objects~\cite{PoseGR2017}. Georgakis \etal train model with RGB-D images and test on trained category unseen instances \cite{MatchingRI2018}. 
% In this work, as we evaluate object pose estimation on novel categories without any preliminary information about test object reference pose during training, pose estimation with given novel object shape is the reasonable setting. Inspired by recent works using object shape, we explore category-free pose estimation with different shape representations and achieved desirable results.   
In this work, we propose to merge the two lines of work described above. We cast pose estimation as a prediction problem, similar to deep learning methods that do not explicitly leverage viewpoint information. However, we condition our network on the 3D model of a single instance, represented either by a set of views or a point cloud, allowing our network to rely on the exact 3D model, similarly to the feature and template matching methods. To the best of our knowledge, we are the first to combine image and shape information as input to a network to estimate the relative orientation of the depicted object with respect to the shape.

%-------------------------------------------------------------------------
\section{Network Architecture and Training}
\label{sec:method}

\begin{figure}
    \centering
	\begin{tabular}{c@{~~}c}
    \bmvaHangBox{\hspace{-2mm}\includegraphics[height=.245\linewidth]{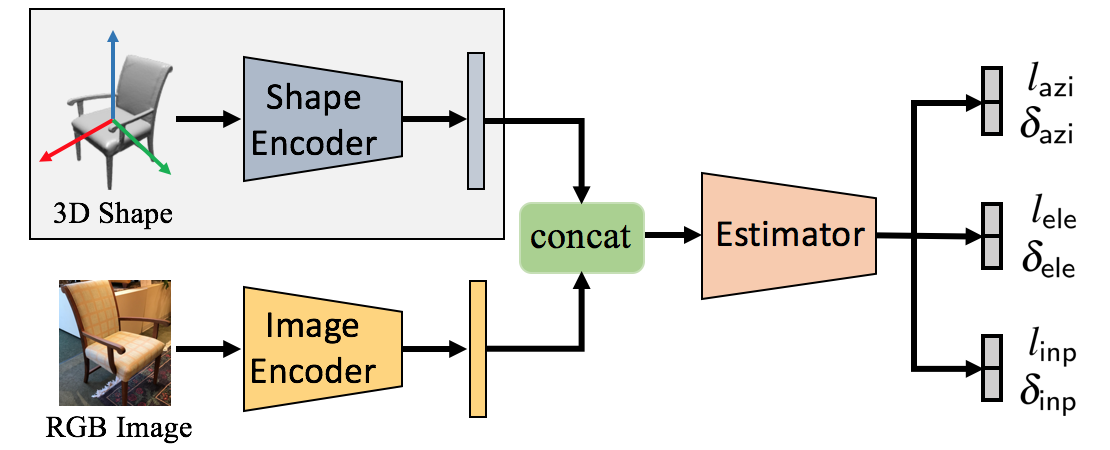}}&
    \bmvaHangBox{\includegraphics[height=.23\linewidth]{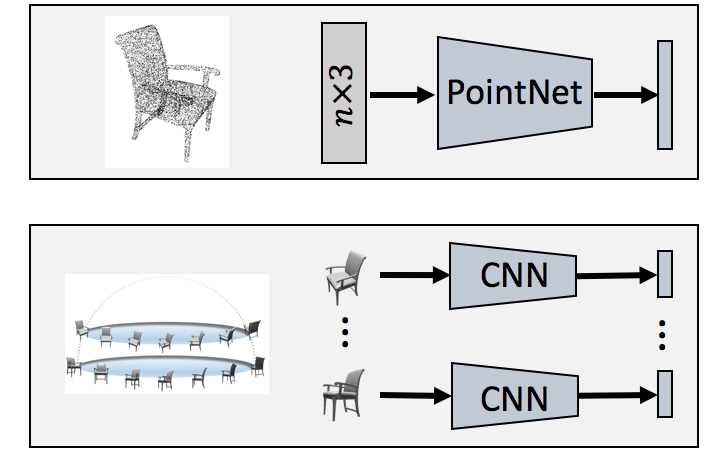}}\\
    (a) Our pose estimation approach & {(b) Two possible shape encoders}
    \end{tabular}
    \vspace{-2mm}
	\caption{Overview of our method. (a) Given an RGB image of an object and its 3D shape, we use two encoders to extract features from each input,
	% and concatenate them into a global feature, then
	then estimate the orientation of the pictured object w.r.t.\ the shape using a classification-and-regression approach, predicting probabilities of angle bins $l$ and bin offsets $\delta$% within each bin
	, for azimuth, elevation and in-plane rotation. %offsets $\delta$ within angular bins $l$, for azimuth, elevation and in-plane rotation. 
	(b) For shape encoding, we encode a point cloud sampled on the object with PointNet (top), or we rendered images around the object and use a CNN to extract the features (bottom).}
	\vspace{-1mm}
	\label{fig:overview}
\end{figure}

Our approach consists in extracting deep features from both the image and the shape, and using them jointly to estimate a relative orientation. An overview is shown in Fig.~\ref{fig:overview}. In this section, we present in more details our architecture, our loss function and our training strategy, as well as a data augmentation scheme specifically designed for our approach.

\paragraph{Feature extraction.}

The first part of the network consists of two independent modules: (i)~image feature extraction and (ii)~3D shape feature extraction. For image features, we use a standard CNN, namely ResNet-18~\cite{He2016DeepRL}. For 3D shape features, we experimented with two approaches depicted in Fig.~\ref{fig:overview}(b) which are state-of-the-art 3D shape description networks.

First, we used the point set embedding network PointNet~\cite{Qi2017PointNetDL}, % and PointNet++~\cite{Qi2017PointNetDH}, 
which %. The main idea behind such networks is 
has been successfully used as a point cloud encoder for many tasks~\cite{engelmann2017exploring,atlasnet2018,qi2018frustumpointnet,wang2018dynamichgraphcnn, xu2018pointfusion}. 
% computes high dimensional representations of each point, then average over all points. 

Second, we tried to represent the shape using rendered views, similar to \cite{mvcnn2015}. Virtual cameras are placed around the 3D shape, pointing towards the centroid of the model; the associated rendered images are taken as input by CNNs, sharing weights for all viewpoints, which extract image descriptors; a global feature vector is obtained by concatenation. We considered variants of this architecture using extra input channels for depth and/or surface normal orientation but this did not improve our results significantly. %As in our experiments all alternatives had similar performances, we kept the simplest architecture, i.e., rendered images.
 Ideally, we would consider viewpoints on the whole sphere around the object with any orientation. In practice however, many objects have a strong bias regarding verticality and are generally seen only from the side/top. In our experiments, we thus only considered viewpoints on the top hemisphere
 %. Besides, viewpoints could be set at even or random locations; what matters is that the configuration of viewpoint locations does not change.  In our experiments, for simplicity and to comply with verticality, we choose to 
 and sampled evenly a fixed number of azimuths and elevations.

\paragraph{Orientation estimation.}

The object orientation is estimated from both the image and 3D shape features by a multi-layer perceptron (MLP) with three hidden layers of size 800-400-200. Each fully connected layer is followed by a batch normalization, and a ReLU activation.  
%Once we have features extracted from both the image and the 3D shape, the object orientation is estimated by a multi-layer perceptron (MLP) with three hidden layers of size 800-400-200. Each fully connected layer is followed by a batch normalization, and a ReLU activation.  

As output, we estimate the three Euler angles of the camera, azimuth ($\azi$), elevation ($\ele$) and in-plane rotation ($\inp$), with respect to the shape reference frame.
% TODO: In order to tackle the ambiguities that may exist between different poses, \cite{RenderForCNN2015,ViewpointsKeypoints2015} divide the angles into non-overlaying bins and solve a classification problem, while~\cite{Mahendran2018AMC,UnifiedMVMC2018} use a mixed classification-regression framework to estimate the pose.
Each of these angles $\theta \in \Euler = \{\azi,\ele,\inp\}$ is estimated using a mixed classification-and-regression approach, which computes both angular bin classification scores and offset information within each bin. Concretely, we split each angle $\theta \in \Euler$ uniformly in $L_\theta$ bins. For each $\theta$-bin $l\in \{0,L_\theta-1\}$, the network outputs a probability $\shat{p}_{\theta,l} \in [0,1]$ using a softmax non-linearity on the $\theta$-bin classification scores, and an offset $\shat{\delta}_{\theta,l} \in [{-}1, 1]$ relatively to the center of $\theta$-bin $l$, obtained by a hyperbolic tangent non-linearity. The network thus has $2\times(L_\azi+L_\ele+L_\inp) $ outputs.
% Half of them correspond to predicted offsets within each bin for each of the three angles and are produced by a hyperbolic tangent non-linearity. The other half is divided in three groups corresponding to the different angles, and the predictions within each group are related by a softmax non-linearity. 
%The predicted angle is $\shat{\theta} = L_\theta(\shat{l}_{\theta} + \smash{\frac{1}{2}\shat{\delta}—_{\theta,\shat{l}_\theta}})$ \todo{ this formula is both wrong, hugly, and not easy to write well since the range is different depending on the angle... I (Mathieu) would just remove this last sentence} where 
% $\smash{\shat{l}_{\theta}} = \argmax_l \shat{p}_{\theta}$ is the label of the $\theta$-bin with the highest score.

% We treat our pose estimation problem as a combination of classification and regression problem, where a bin size of 15 degrees is used for angle bin classification and a delta value is estimated within the correct bin for each angle. Thus, the predicted angle is: $\shat{\theta} = (\shat{l}_{\theta} + \delta_{\theta}) \times B$, where $\shat{l}_{\theta}$ is the angle bin label, $\delta_{\theta} \in [0, 1)$ is the delta value and $B$ is the bin size. $\theta$ could be any of the three Euler angles.

\paragraph{Loss function.}

As we combine classification and regression, our network has two types of outputs (probabilities and offsets), that are combined into a single loss $\Loss$ that is the sum of a cross-entropy loss for classification $\Loss_\class$ and Huber loss~\cite{huberloss} for regression $\Loss_\reg$.

More formally, we assume we are given training data $(x_i, s_i, y_i)_{i=1}^N$ consisting of input images $x_i$, associated object shapes $s_i$ and corresponding orientations $y_i = (y_{i,\theta})_{\theta\in\Euler}$. We convert the value of the Euler angles $y_{i,\theta}$ into a bin label $l_{i,\theta}$ encoded as a one-hot vector and relative offsets $\delta_{i,\theta}$ within the bins.  The network parameters are learned by minimizing:
\begin{equation}
    \Loss = \sum_{i=1}^N \ \sum_{\theta \in \Euler} \Loss_\class  \Big(l_{i,\theta}, \shat{p}_{\theta}(x_i,s_i)\Big) + \Loss_\reg \Big(\delta_{i,\theta}, \shat{\delta}_{\theta,l_{i,\theta}}(x_i,s_i)\Big),
\end{equation}
where $\shat{p}_{\theta}(x_i,s_i)$ are the probabilities predicted by the network for angle $\theta\in\Euler$, input image $x_i$ and input shape $s_i$, and $\shat{\delta}_{\theta,l_{i,\theta}}(x_i,s_i)$ the predicted offset within the ground truth bin.% for this same inputs.

\paragraph{Data augmentation.}

% As proposed or used in previous work~\cite{RenderForCNN2015,3DPR2017,pix3d2018}, 
We perform standard data augmentation on the input images: horizontal flip, 2D bounding box jittering, color jittering.

In addition, we introduce a new data augmentation, specific to our approach, designed to avoid the network to overfit the 3D model orientation, which is usually consistent in training data since most models are aligned. On the contrary, we want our network to be category-agnostic and to always predict the pose of the object with respect to the reference 3D model. We thus add random rotations to the input shapes, and modify the orientation labels accordingly.
% Regarding the input shapes, we enforce the insensitivity to their given reference frames by randomly rotating the shapes, adjusting the expected poses accordingly. Again, we observe a strong verticality bias in existing datasets, with a wide range of azimuths as opposed to little variation in elevations and in-plane rotations.  
In our experiments, we restrict our rotations to azimuth changes, again because of the strong verticality bias in the benchmarks, but could theoretically apply it to all angles. Because of objects with symmetries, typically at $90\degree$ or $180\degree$, we also restrict azimuthal randomization to a uniform sampling in $[{-}45\degree,45\degree]$, which allows to keep the $0\degree$ bias of the annotations. % This randomization is used when extracting 3D features as well as 2D multiview features.
See supplementary material for details and parameter study.
% Other sampling ranges are evaluated in the Supplementary, where we found that this setting gives the best result.

% In other words, we can exclude the 3D feature extraction (blue part in Fig~\ref{fig:overview}) and solve the classification and regression problem by using only the image descriptor given by the 2D image (yellow part in Fig~\ref{fig:overview}). This involves an implicit recognition for different object categories and the orientation.

\paragraph{Implementation details.}
 For all our experiments, we set the batch size as 16 and trained our network using the Adam optimizer~\cite{Kingma2015AdamAM} with a learning rate of $10^{-4}$ for 100 epochs then $10^{-5}$ for an additional 100 epochs. Compared to a shape-less baseline method, the training of our method with the shape encoded from 12 rendered views is about 8 times slower, on a TITAN X GPU.

%-------------------------------------------------------------------------
\section{Experiments}
\label{sec:exps}
Given an RGB image of an object and a 3D model of that object, our method estimates its 3D orientation in the image. %We used It is typically to be used after objects are detected on images with their bounding boxes~\cite{Ren2015FasterRT}, from which object images can be cropped. Yet, in our experiments, to separate detection and orientation issues, we only consider the ``oracle'' ground-truth bounding boxes provided with the various datasets.
In this section, we first give an overview of the datasets we used, and explain our baseline methods. We then evaluate our method in two test scenarios: 
object belonging to a category known at training time, or unknown. %\todo{ Mathieu: is "Discussion and limitations" really necessary? Can we not justdo it in both sections: Finally, we discuss our results and limitations of our approach.}

% analyze and evaluate our approach. We first present the main datasets we used, including a synthetic training set generated using 3D models of the ShapeNet database~\cite{shapenet2015} and 2D images from SUN397 database~\cite{SUN2010}. Second, we provide detailed results and analysis for the task of category-agnostic pose estimation on supervised categories and novel categories. Finally, we present quantitative and qualitative results for pose estimation on unseen objects possessing arbitrary shapes.

\paragraph{Datasets.}

We experimented with four main datasets.  Pascal3D+~\cite{pascal3d2014}, ObjectNet3D~\cite{object3d2016} and Pix3D~\cite{pix3d2018} feature various objects in various environments, allowing benchmarks for object pose estimation in the wild. On the contrary, LINEMOD~\cite{hinterstoisser2012LINEMOD} focuses on few objects with little environment variations, targeting robotic manipulation. Pascal3D+ and ObjectNet3D only provide approximate models and rough alignments while Pix3D and LINEMOD offer exact models and pixelwise alignments. We also used ShapeNetCore~\cite{shapenet2015} for training on synthetic data, with SUN397 backgrounds~\cite{SUN2010}, and tested on Pix3D and LINEMOD. 

Unless otherwise stated, ground-truth bounding boxes are used in all experiments. 
We compute the most common metrics used with each dataset: $\Accpisix$ is the percentage of estimations with rotation error less than $30\degree$; \MedErr\ is the median angular error (\textdegree);  
\ADD\ is the percentage of estimations for which the mean distance of the estimated 3D model points to the ground truth is smaller than 10\% of the object diameter; \ADDS\ is a variant of {\ADD} used for symmetric objects where the average is computed on the closest point distance. More details on the datasets and metrics are given in the supplementary material.

\paragraph{Baselines.}

A natural baseline is to use the same architecture, data and training strategy as for our approach, but without using the 3D shape of the object. This is reported as `Baseline' in our tables, and corresponds to the network of Fig.~\ref{fig:overview} without the shape encoder shown in light blue. % \todo{describe depending on the figure}, when classification and regression information are estimated solely from the input image features. 
We also report a second baseline, aiming at evaluating the importance of the precision of the 3D model for our approach to work. We used exactly our approach, but at testing time we replaced the 3D shape of the object in the test image by a random 3D shape of the same category. This is reported as `Ours (RS)' in the tables. %\todo{add random 3D model baseline if we add it everywhere}

%-------------------------------------------------------------------------
\subsection{Pose estimation on supervised categories}

We first evaluate our method in case the categories of tested objects are covered by training data. We show that leveraging the 3D model of the object clearly improves pose estimation.

\begin{table}[p]
	\centering
	\bgroup
	\scriptsize \addtolength{\tabcolsep}{-5pt}
		\begin{tabular}{l c c c c c c c c c c c c c c c c c c c c |>{\columncolor{HL}}c}
		\toprule
		\textbf{ObjectNet3D}~%\cite{object3d2016}
		& bed & bcase & calc & cphone & comp & door & cabi & guit & iron & knife & micro & pen & pot & rifle & shoe & slipper & stove & toilet & tub & wchair & mean \\ % total \\
		% \#\,images & 456 & 200 & 199 & 185 & 123 & 284 & 232 & 103 & 124 & 265 & 208 & 284 & 270 & 137 & 180 & 211 & 291 & 115 & 107 & 255 & 4229 \\
		% \#\,models & 10 & 8 & 5 & 11 & 12 & 14 & 8 & 5 & 5 & 8 & 6 & 7 & 7 & 8 & 10 & 6 & 6 & 7 & 9 & 5 & 157 \\
		\midrule & \multicolumn{21}{c}{\textbf{category-specific networks/branches --- test on supervised categories} \ $\bm{(\Accpisixb)}$}\\ \midrule
		
		Xiang~\cite{object3d2016}* & 61 & 85 & 93 & 60 & 78 & 90 & 76 & 75 & 17 & 23 & 87 & 33 & 77 & 33 & 57 & 22 & 88 & 81 & 63 & 50 & 62 \\
		
		\midrule & \multicolumn{21}{c}{\textbf{category-agnostic network --- test on supervised categories} \ $\bm{(\Accpisixb)}$} \\ % & \textbf{mean} \\
		\midrule
		
		Zhou~\cite{starmap2018} & 73 & 78 & 91 & 57 & 82 & -- & 84 & 73 & 3 & 18 & 94 & 13 & 56 & 4 & -- & 12 & 87 & 71 & 51 & 60 & 56 \\
		%Baseline(cls) & 70 & 86 & 89 & 60 & 85 & 88 & 85 & 66 & 38 & 22 & 92 & 41 & 68 & 27 & 54 & 29 & 87 & 65 & 56 & 57 & 63 \\
		Baseline & 70 & 89 & 90 & 55 & 87 & 91 & 88 & 62 & 29 & 20 & 93 & 43 & 76 & 26 & 58 & 30 & 91 & 68 & 51 & 55 & 64 \\
		%Ours(cls) & 72 & 87 & 93 & \textbf{68} & 87 & 91 & \textbf{87} & 65 & 34 & 26 & 93 & 48 & 80 & 34 & 62 & \textbf{42} & 91 & 77 & 74 & 51 & 68 \\
		Ours(PC) & \textbf{83} & \textbf{92} & 95 & 58 & 82 & 87 & \textbf{91} & 67 & 43 & \textbf{36} & 94 & 53 & 81 & 39 & 45 & 35 & 91 & 80 & 65 & 56 & 69 \\
		Ours(MV,RS) & 74 & 89 & 91 & 62 & 81 & 90 & 88 & 71 & 41 & 28 & 94 & 50 & 70 & 37 & 57 & 38 & 89 & 81 & 60 & 60 & 68 \\
		Ours(MV) & 82 & 90 & \textbf{96} & \textbf{65} & \textbf{93} & \textbf{97} & 89 & \textbf{75} & \textbf{52} & 32 & \textbf{95} & \textbf{54} & \textbf{82} & \textbf{45} & \textbf{67} & \textbf{46} & \textbf{95} & \textbf{82} & \textbf{67} & \textbf{66} & \textbf{73} \\
		\midrule & \multicolumn{21}{c}{\textbf{category-agnostic network --- test on novel categories} \ $\bm{(\Accpisixb)}$} \\ % & \textbf{mean}\\
		\midrule
		Zhou~\cite{starmap2018} & 37 & 69 & 19 & 52 & 73 & -- & 78 & 61 & 2 & 9 & 88 & 12 & 51 & 0 & -- & 11 & 82 & 41 & 49 & 14 & 42 \\
		%Baseline(cls) & 52 & 75 & 52 & 42 & 75 & 77 & 82 & 25 & 2 & 22 & 88 & 43 & 64 & 1 & 30 & 20 & 82 & 42 & 49 & 36 & 48 \\
		Baseline & 56 & 79 & 26 & 53 & 77 & 86 & 83 & 51 & 4 & 16 & 90 & 42 & 65 & 2 & 34 & 22 & 86 & 43 & 50 & 35 & 50 \\
		%Ours(cls) & 62 & 88 & 82 & \textbf{62} & 87 & 91 & \textbf{88} & \textbf{82} & \textbf{5} & 19 & \textbf{95} & 47 & 76 & 7 & \textbf{51} & 26 & 87 & \textbf{51} & 62 & \textbf{40} & 61 \\
		Ours(PC) & 63 & 85 & 84 & 51 & \textbf{85} & 83 & 83 & 61 & \textbf{9} & \textbf{35} & 92 & 44 & \textbf{80} & 8 & 39 & 20 & 87 & 56 & \textbf{71} & \textbf{39} & 59 \\
		Ours(MV,RS) & 60 & 88 & 84 & 60 & 76 & 91 & 82 & 61 & 2 & 26 & 90 & 46 & 73 & 13 & 45 & 28 & 79 & 59 & 61 & 36 & 58\\
		Ours(MV) & \textbf{65} & \textbf{90} & \textbf{88} & \textbf{65} & 84 & \textbf{93} & \textbf{84} & \textbf{67} & 2 & 29 & \textbf{94} & \textbf{47} & 79 & \textbf{15} & \textbf{54} & \textbf{32} & \textbf{89} & \textbf{61} & 68 & \textbf{39} & \textbf{62} \\ % \midrule
		\bottomrule	
		\multicolumn{22}{c}{[images: 90,127, in the wild | objects: 201,888 | categories: 100 | 3D models: 791, approx.\ | align.: rough]} \\
	\end{tabular}
	\vspace{-2mm}
	\caption{Pose estimation on ObjectNet3D~\cite{object3d2016}. % (90,127 images in the wild, 201,888 objects in 100 categories, 791 approximate 3D models, rough alignments). 
	Train and test are on the same data as~\cite{starmap2018}; % the \todo{XXX} objects (supervised case) or \todo{XXX} objects (agnostic case) having also keypoints annotations. 
	for experiments on novel categories, training is on 80 categories and test is on the other 20. %, as~\cite{starmap2018}. %All tests are on the same \todo{XXX} images. 
	*~Trained jointly for detection and pose estimation, tested using estimated bounding boxes.
	% Accuracy is measured as $\Accpisix$ (3D rotation error < $30\degree$).
	}
	\label{tab:object3d}
		\egroup
%\end{table}
\vspace{5.5mm}
%\begin{table*}
%	\centering
    \bgroup
	\scriptsize \addtolength{\tabcolsep}{-4.9pt}
	\begin{tabular}{l c c c c c c c c c c c c |>{\columncolor{HL}}c|cccccccccccc|>{\columncolor{HL}}c}
		\toprule
		\textbf{Pascal3D+}~\cite{pascal3d2014} &
		\rotatebox[origin=c]{90}{aero} & \rotatebox[origin=c]{90}{bike} & \rotatebox[origin=c]{90}{boat} & \rotatebox[origin=c]{90}{bottle} & \rotatebox[origin=c]{90}{bus} & \rotatebox[origin=c]{90}{car} & \rotatebox[origin=c]{90}{chair} & \rotatebox[origin=c]{90}{dtable} & \rotatebox[origin=c]{90}{mbike} & \rotatebox[origin=c]{90}{sofa} & \rotatebox[origin=c]{90}{train} & \rotatebox[origin=c]{90}{tv} & \rotatebox[origin=c]{90}{mean} & \rotatebox[origin=c]{90}{aero} & \rotatebox[origin=c]{90}{bike} & \rotatebox[origin=c]{90}{boat} & \rotatebox[origin=c]{90}{bottle} & \rotatebox[origin=c]{90}{bus} & \rotatebox[origin=c]{90}{car} & \rotatebox[origin=c]{90}{chair} & \rotatebox[origin=c]{90}{dtable} & \rotatebox[origin=c]{90}{mbike} & \rotatebox[origin=c]{90}{sofa} & \rotatebox[origin=c]{90}{train} & \rotatebox[origin=c]{90}{tv} & \rotatebox[origin=c]{90}{mean} \\ % \textbf{total} \\
		%& aero & bike & boat & bottle & bus & car & chair & dtable & mbike & sofa & train & tv & mean \\ % \textbf{total} \\
		%\#\,images & 275 & 118 & 232 & 251 & 154 & 308 & 244 & 21 & 136 & 39 & 113 & 222 & 2113 \\
		%\#\,models & 8 & 6 & 6 & 8 & 6 & 10 & 10 & 6 & 5 & 6 & 4 & 4 & 79 \\
		\midrule 
		\multicolumn{14}{l|}{\textbf{Categ.-specific branches, supervised categ.} \hspace{3mm} $\Accpisix$ (\%)} & \multicolumn{13}{c}{~\MedErr~(degrees)} \\ % & \textbf{mean} \\ 
		%& \multicolumn{26}{c}{\textbf{category-specific branches --- test on supervised categories} \ $(\,\Accpisix ~|~ \MedErr)$} \\% & \textbf{mean} \\ 
		\midrule
		Tulsiani~\cite{ViewpointsKeypoints2015}* & 81 & 77 & 59 & 93 & \textbf{98} & 89 & 80 & 62 & \textbf{88} & 82 & 80 & 80 & 80.75 & 13.8 & 17.7 & 21.3 & 12.9 & 5.8 & 9.1 & 14.8 & 15.2 & 14.7 & 13.7 & 8.7 & 15.4 & 13.6 \\
		Su~\cite{RenderForCNN2015}$\dag$ & 74 & \textbf{83} & 52 & 91 & 91 & 88 & \textbf{86} & \textbf{73} & 78 & \textbf{90} & \textbf{86} & \textbf{92} & 82.00 & 15.4 & 14.8 & 25.6 & 9.3 & 3.6 & 6.0 & \textbf{9.7} & \textbf{10.8} & 16.7 & \textbf{9.5} & \textbf{6.1} & 12.6 & 11.7 \\
		Mousavian~\cite{3DBboxMousavian2017} & 78 & \textbf{83} & 57 & 93 & 94 & 90 & 80 & 68 & 86 & 82 & 82 & 85 & 81.03 & 13.6 & \textbf{12.5} & 22.8 &\textbf{8.3} & 3.1 & 5.8 & 11.9 & 12.5 & 12.3 & 12.8 & 6.3 & 11.9 & 11.1 \\
		Pavlakos~\cite{pavlakos20176}* & 81 & 78 & 44 & 79 & 96 & 90 & 80 & -- & -- & 74 & 79 & 66 & -- & \textbf{8.0} & 13.4 & 40.7 & 11.7 & \textbf{2.0} & 5.5 & 10.4 & -- & -- & 9.6 & 8.3 & 32.9 & --\\
		Grabner~\cite{3Dpose3Dmodel2018} & \textbf{83} & 82 & \textbf{64} & \textbf{95} & 97 & \textbf{94} & 80 & 71 & \textbf{88} & 87 & 80 & 86 & \textbf{83.92} & 10.0 & 15.6 & \textbf{19.1} & 8.6 & 3.3 & \textbf{5.1} & 13.7 & 11.8 & \textbf{12.2} & 13.5 & 6.7 & \textbf{11.0} & \textbf{10.9} \\
		\midrule 
		%& \multicolumn{26}{c}{\textbf{category-agnostic network --- test on supervised categories} \ $(\,\Accpisix ~|~ \MedErr)$} \\ % & \textbf{mean} \\ 
		\multicolumn{14}{l|}{\textbf{Categ.-agnostic network, supervised categ.} \hspace{3mm} $\Accpisix$ (\%)} & \multicolumn{13}{c}{~\MedErr ~(degrees)} \\ % & \textbf{mean} \\ 
		\midrule	
		Grabner~\cite{3Dpose3Dmodel2018} & 80 & 82 & 57 & 90 & \textbf{97} & \textbf{94} & 72 & \textbf{67} & \textbf{90} & 80 & \textbf{82} & \textbf{85} & 81.33 & 10.9 & \textbf{12.2} & 23.4 & 9.3 & 3.4 & 5.2 & 15.9 & 16.2 & 12.2 & 11.6 & 6.3 & 11.2 & 11.5 \\
		Zhou~\cite{starmap2018}* & \textbf{82} & \textbf{86} & 50 & 92 & \textbf{97} & 92 & \textbf{79} & 62 & 88 & \textbf{92} & 77 & 83 & 81.67 & \textbf{10.1} & 14.5 & 30.0 & 9.1 & 3.1 & 6.5 & 11.0 & 23.7 & 14.1 & 11.1 & 7.4 & 13.0 & 12.8\\
		%Baseline(cls) & 79 & 72 & 50 & 90 & 96 & 87 & 73 & 65 & 88 & 75 & \textbf{83} & 78 & 78.00 \\
		Baseline & 77 & 74 & 54 & 91 & \textbf{97} & 89 & 74 & 52 & 85 & 80 & 79 & 77 & 77.42 & 13.0 & 18.2 & 27.3 & 11.5 & 6.8 & 8.1 & 15.4 & 20.1 & 14.7 & 13.2 & 10.2 & 14.7 & 14.4 \\
		%Ours(cls) & \textbf{85} & 80 & 56 & 91 & 95 & 92 & \textbf{82} & 62 & 87 & 90 & 80 & 81 & 81.75 \\
		Ours(MV,RS) & 79 & 81 & 49 & 91 & 96 & 89 & 78 & 53 & \textbf{90} & 88 & 80 & 77 & 79.25 & 11.6 & 15.5 & 30.9 & 8.2 & 3.6 & 6.0 & 13.8 & 22.8 & 13.1 & 11.1 & 6.0 & 15.0 & 13.1 \\
		Ours(MV) & 81 & 83 & \textbf{60} & \textbf{93} & \textbf{97} & 91 & \textbf{79} & \textbf{67} & \textbf{90} & 90 & 81 & 79 & \textbf{82.66} & 10.5 & 13.7 & \textbf{21.0} & \textbf{7.7} & \textbf{3.0} & \textbf{5.0} & \textbf{10.9} & \textbf{11.9} & \textbf{11.8} & \textbf{9.1} & \textbf{5.4} & \textbf{10.3} & \textbf{10.0} \\
		%\midrule
		\bottomrule	
		\multicolumn{27}{c}{[images: 30,889, in the wild | objects: 36,292 | categories: 12 | 3D models: 79, approx.\ | align.: rough]} \\
	\end{tabular}
	\vspace{1mm}
	% \caption{Pose estimation on Pascal3D+ (train \& test). *Trained on ShapeNet renderings too.}
	\caption{
	%\todo{why are they 2 numbers on the left and 3 on the right? what about 2 numbers for everybody except 3 for the means?}
	Pose estimation on Pascal3D+~\cite{pascal3d2014}. % (30,899 images in the wild, 36,292 objects in 12 categories, 79 approximate 3D models, rough alignments).
	%Trained on ImageNet-trainval and Pascal-train, tested on 2113 non-occluded and non-truncated objects of Pascal-val. 
	*~Trained using keypoints. $\dag$~Not trained on ImageNet data but trained on ShapeNet renderings.} % Accuracy measured as $\Accpisix$ (3D rotation error < $30\degree$).}
	\label{tab:pascal3d}

	\egroup
%\end{table*}
\vspace{5mm}
%\begin{table}
%	\centering
    \bgroup
	\scriptsize \addtolength{\tabcolsep}{-3pt}
	\begin{tabular}[t]{l c c c c c c c c c|>{\columncolor{HL}}c}%||c c}
		\toprule
		\textbf{Pix3D}~\cite{pix3d2018}
		& tool & misc & \!\!\!\!\!bcase\!\! & \!\!wdrobe\!\!\!\! & desk & bed & table & sofa & chair & mean %& \multicolumn{2}{c}{chair} 
		\\
		% \#\,images & 46 & 61 & 130 & 166 & 297 & 394 & 738 & 1092 & 2894 & 5711 \\
		% \#\,models & 8 & 13 & 17 & 10 & 23 & 20 & 63 & 20 & 221 & 395 \\
		\midrule & \multicolumn{10}{c}{\textbf{category-specific networks --- tested on supervised categories} \ $\bm{(\Accpisixb)}$} % & azi & ele 
		\\
		\midrule
		Georgakis~\cite{MatchingRI2018} & - & - & - & - & \textbf{34.9} & \textbf{50.8} & - & - & \textbf{31.2} & - \\ % & - & - \\
        %Su~\cite{RenderForCNN2015} &-&-&-&-&-&-&-&-&-&-& 40 & 37 \\
        % Sun~\cite{pix3d2018} &-&-&-&-&-&-&-&-&-&-& \textbf{49} & \textbf{61} \\
        \midrule
		& \multicolumn{10}{c}{\textbf{category-agnostic network --- tested on supervised categories}  \ $\bm{(\Accpisixb)}$} \\ % & \textbf{mean} 
		\midrule
		Baseline & 2.2 & 9.8 & 10.8 & 0.6 & 30.0 & 36.8 & 17.3 & 63.8 & 43.6 & 23.9 \\ % & 51 & 64 \\  
		Ours(MV,RS)\!\!\! & 4.1 & 3.6 & 22.8 & 9.5 & 52.8 & 50.1 & 30.8 & 66.3 & 44.5 & 31.6 \\ % & - & - \\
		Ours(MV) & \textbf{6.5} & \textbf{19.7} & \textbf{34.6} & \textbf{10.2} & \textbf{56.6} & \textbf{59.8} & \textbf{40.8} & \textbf{70.0} & \textbf{52.4} & \textbf{38.9} \\ % & \textbf{54} & \textbf{65} \\

		\midrule & \multicolumn{10}{c}{\textbf{category-agnostic network --- tested on novel categories} \ $\bm{(\Accpisixb)}$} \\ % & \textbf{mean} \\
		\midrule
		Baseline & 2.2 & \textbf{13.1} & 5.4 & 0.6 & 30.3 & 19.6 & 14.9 & 11.9 & 28.0 & 14.0 \\ % & - & - \\
		Ours(MV,RS)\!\!\! & 3.0 & 5.9 & 4.5 & 5.2 & 44.7 & 31.5 & 24.1 & 48.5 & 33.9 & 22.4 \\ % & - & - \\
		Ours(MV) & \textbf{10.9} & \textbf{13.1} & \textbf{22.3} & \textbf{6.6} & \textbf{52.0} & \textbf{55.3} & \textbf{35.6} &  \textbf{64.6} & \textbf{35.8} & \textbf{32.9} \\ % & - & - \\
		%\midrule
		\bottomrule
		\multicolumn{11}{c}{[images: 10,069, in the wild | objects: 10,069 | categ.: 9 | 3D models: 395, exact\ | align.: pixel]} \\
	\end{tabular}
	\quad\begin{tabular}[t]{lcc}
		\toprule
		\textbf{Pix3D}~\cite{pix3d2018} &\multicolumn{2}{c}{chair} \\
		\midrule
		\multicolumn{3}{c}{\textbf{categ.-specific, supervised}} \\
		\midrule
		\#\,bins & 24 & 12 \\
		\hspace{1mm}\rlap{(\% correct)} & azim. & elev. \\
		\midrule
		Su~\cite{RenderForCNN2015}  & 40 & 37\\
		Sun~\cite{pix3d2018} & 49 & 61 \\
		Baseline & 51 & 64 \\
		%Ours(MV,RS) & 34 & 44 \\
		Ours(MV) & \textbf{54} & \textbf{65} \\
		\bottomrule
	\end{tabular}
	\vspace{2mm}
	\caption{Pose estimation on Pix3D~\cite{pix3d2018}. % (10,069 images in the wild, 10,069 selected objects in 9 categories, 395 ``exact'' 3D models, pixelwise alignments). %Test on 5,711 images/objects.
	% Accuracy is measured as $\Accpisix$ (3D rotation error < $30\degree$).
	Right table compares to \cite{RenderForCNN2015,pix3d2018}, that only test bin success on 2 angles (24 azimuth bins and 12 elevation bins). 
	% \todo{explain the metrics on both tables}
	}
	\label{tab:pix3d_all}
	\egroup
\end{table}

% \paragraph{Comparison and analysis on other datasets.} 

%\begin{comment}
%\begin{wraptable}{r}{6cm}
%\centering
%\scriptsize
%		\label{tab:pix3d_chair}
%		\begin{tabular}{l|cc}
%			\toprule
%			& Azimuth & Elevation \\ \midrule
%			Su~\cite{RenderForCNN2015}  & 0.40 & 0.37\\
%			Pix3D~\cite{pix3d2018} & 0.49 & 0.61 \\
%			Baseline & 0.51 & 0.64 \\
%			Ours(MV) & \textbf{0.54} & \textbf{0.65} \\ \midrule
%			\bottomrule
%		\end{tabular}
%		\caption{Results on chairs of Pix3D~\cite{pix3d2018}, shown in top-1 classification accuracy of 24 views for azimuth and 12 views for elevation.}
%		\vspace{-2mm}
%\end{wraptable}
%\end{comment}

% \paragraph{Validation on Pascal3D+.}

We evaluate our method on ObjectNet3D, which has the largest variety of object categories, 3D models and images. We report the results in Table~\ref{tab:object3d} (top). First, an important result is that using the 3D model information, whether via %in the network,  through 
a point cloud or rendered views, % encoder 
provides a very clear boost of the performance, which validates our approach. Second, % one can see that
results using rendered multiple views (MV) to represent the 3D model outperform the point-cloud-based (PC) representation~\cite{Qi2017PointNetDL}. We thus only evaluated Ours(MV) %the rendered-view based model 
in the rest of this section. 
Third, testing the network with a random shape (RS) in the category instead of the ground truth shape, implicitly providing class information without providing fine-grained 3D information, leads to results better than the baseline but worst than using the ground truth model, demonstrating our method ability to exploit fine-grained 3D information. Finally, we found that even our baseline model already outperformed StarMap~\cite{starmap2018}, mainly because of five categories (iron, knife, pen, rifle, slipper) on which StarMap completely fails, likely because a keypoint-based method is not adapted for small and narrow objects.
% \todo{comment random 3D model baseline}
%This dataset contains objects coming from 100 rigid categories while Pascal3D+ contains only 12 categories. 
%Thus, it is easier to learn common inter-category information with such richer category samples for methods based on learning from images.
% We find that our baseline model has already outperformed StarMap~\cite{starmap2018} when their approach failed (with accuracy lower than 20\%) on certain categories (Iron, Knife, Pen, Rifle, Slipper). One explanation is that these 5 failed categories are small and narrow objects, which are difficult for StarMap~\cite{starmap2018} as they relied on keypoint detection. This indicates that methods based on keypoint detection and PnP algorithm are not robust enough in the case where the object instances are small and textureless. When the 3D model information is inserted in the network, a clear performance improvement is achieved (74\% $Acc_{\frac{\pi}{6}}$ compared to 56\%).

We then evaluate our approach on the standard Pascal3D+ dataset~\cite{pascal3d2014}. Results are shown in Table~\ref{tab:pascal3d} (top). Interestingly, while our baseline is far below state-of-the-art results, adding our shape analysis network provides again a very clear improvement, with results on par with the best category-specific approaches, and outperforming category agnostic methods. This is especially impressive considering the fact that the 3D models provided in Pascal3D+ are only extremely coarse approximations of the real 3D models. Again, as can be expected, using a random model from the same category provides intermediary results between the model-less baseline and using the actual 3D model.

Finally, we report results on Pix3D in Table~\ref{tab:pix3d_all} (top). Similar to the other methods, our model was purely trained on synthetic data and tested on real data, without any fine-tuning. Again, we can observe that adding 3D shape information brings a large performance boost, from $23.9\%$ to $36\%$ $\Accpisix$.
%and our category-agnostic approach achieved a better accuracy compared to the results of~\cite{MatchingRI2018} who trained a separate network for each category. Describe Table 2 here. 
Note that our method clearly improves even over category-specific baselines. We believe it is due to the much higher quality of the 3D models provided on Pix3D compared to ObjectNet3D and Pascal3D+. This hypothesis is supported by the fact that our results are much worse when a random model of the same category is provided.
% To validate that our category-free approach is competitive with other methods for pose estimation on different categories, we trained our network using the training data of Pascal3D+ and evaluated it on the non-occluded and non-truncated validation set using the ground truth object bounding box. The comparison with the state-of-the-art methods is shown in Table~\ref{tab:pascal3d}. We can first observe that our method outperforms others under the category-agnostic setting, and a large margin is gained compared to the baseline model. We believe that adding 3D model in the network can help instance pose estimation when only one branch is used for all object categories. Moreover, our category-free approach  

These state-of-the-art results on the three standard datasets are thus consistent and validate (i) that using the 3D models provides a clear improvement (comparison to `Baseline'), and (ii) that our approach is able to leverage the fine-grained 3D information from the 3D model (comparison to estimating with a random shape `RS' in the category). %%Besides, we obtain a very clear improvement over the state of the art both on the ObjectNet3D and Pix3D datasets.

%-------------------------------------------------------------------------
\subsection{Pose estimation on novel categories}

We now focus on the generalization to unseen categories, which is the main focus of our method. We first discuss results on ObjectNet3D and Pix3D. We then show qualitative results on ImageNet horses images and quantitative results on the very different LINEMOD dataset.
% In this part, we focus more specifically on pose estimation for unseen objects and show quantitative and qualitative results on various datasets, including ObjectNet3D, Pix3D and LINEMOD which opens a direct application to the robotic manipulation with texture-less objects in cluttered scenes.

% \paragraph{Quantitative results.} 
% We first compare the performance gap between including and withholding the 20 novel categories of ObjectNet3D during training. 
Our results when testing on new categories from ObjectNet3D are shown in Table~\ref{tab:object3d} (bottom). We use the same split between 80 training and 20 testing categories as~\cite{starmap2018}. 
As expected, the accuracy decreases for all methods when supervision is not provided on these latter categories. The fact that the Baseline performances are still much better than chance is accounted by the presence of similar categories is the training set. 
%explained by the fact that similar categories are present in the training set. 
The advantage of our method is however even more pronounced than in the supervised case, and our multi-view approach (MV) still outperforms the point cloud (PC) approach by a small margin. Similarly, we removed from our ShapeNet~\cite{shapenet2015} synthetic training set the categories present in Pix3D, and reported in Table~\ref{tab:pix3d_all} (bottom) the results on Pix3D. Again, the accuracy drops for all methods, but the benefit from using the ground-truth 3D model increases.

% , especially for some categories (Iron, Rifle, Toilet, Wheelchair), the performance gap is larger than 20\%. One explanation is that these instances possess a very different appearance compared to the annotated CAD model, which makes it difficult since our method is parameterized on the accurate 3D shape. Another reason for this could be the imprecision of the pose annotation in ObjectNet3D when the image is blurred. Furthermore, by scrutinizing through the non-occluded and non-truncated validation set, we found that some mislabeled instances were actually truncated or occluded. However, even with the approximate 2D-3D alignment, our method outperforms the state-of-the-art method StarMap~\cite{starmap2018} by a large margin (62\% $Acc_{\frac{\pi}{6}}$ compared to 42\%) on the 20 novel categories and the performance gap is smaller than 10\% for more than half of the 20 categories. This can also be observed in the results on Pix3D (see Table~\ref{tab:pix3d_all}), where the average performance gap between including and withholding the testing categories is smaller than 10\%. 

% \paragraph{Invariant to the shape texture.} 

% \paragraph{Coupling with Pose Refinement.}

\smallskip

In both ObjectNet and Pix3D experiments, the test categories were novel but still similar to the training ones. We now focus on evaluating our network, trained using synthetic images generated from man-made shapes from ShapeNetCore~\cite{shapenet2015}, on completely different objects.

\begin{figure}
	\centering
	\includegraphics[width=1.0\linewidth]{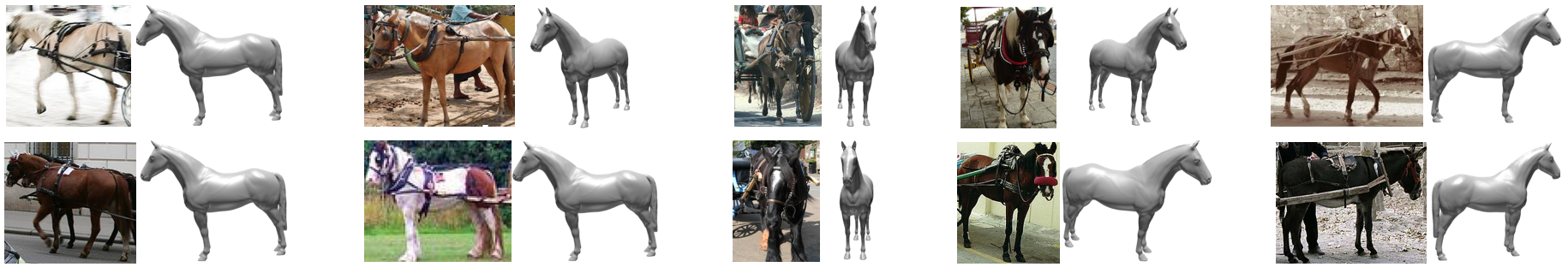}
	\vspace{-6mm}
	\caption{Visual results of pose estimation on horse images from ImageNet~\cite{imagenet_2009} using models from Free3D~\cite{free3d}. We rank the prediction for each orientation bin by the network prediction and show the first (best) results for various poses. %\todo{I like it, can you show more poses? can we show the to 10 for each bin in sup mat? Does the same trick work for other categories?}
	}
	\label{fig:visual_results}
\end{figure}

We first obtain qualitative results by using a fixed 3D model of horse from an online model repository~\cite{free3d} to estimate the pose of horses in ImageNet images. Indeed, compared to other animals, horses have more limited deformations. While this of course does not work for all images, the images for which the network provides the highest confidence are impressively good. On Figure~\ref{fig:visual_results}, we show the most confident images for different poses, and we provide more results in the supplementary material. Note the very strong appearance gap between the rendered 3D models and the test images.

% In all our experiments, we do not require the object in the RGB image to share the same texture as the 3D shape. This can be observed in Figure~\ref{fig:visual_results}, where we estimate the pose of horse images with different textures and poses coming from ImageNet dataset. 

Finally, to further validate our network generalization ability, we evaluate it on the texture-less objects of LINEMOD~\cite{hinterstoisser2012LINEMOD}, as reported in Table~\ref{tab:LINEMOD}. This dataset focuses on very accurate alignment, and most approaches propose to first estimate a coarse alignment and then to refine it with a specific method. Our method provides a coarse alignment, and we complement it using the recent DeepIM~\cite{li2018deepim} refinement approach. Our method yields results below the state of the art, but they are nevertheless very impressive. Indeed, our network has never seen objects any similar the LINEMOD 3D models during training, while all the other baselines have been trained specifically for each object instance on real training images, except SSD-6D~\cite{SSD6D2017} which uses the exact 3D model but no real image and for which coarse alignment performances are very low. Our method is thus very different from all the baselines in that it does not assume the test object to be available at training time, which we think is a much more realistic scenario for robotics applications. We actually believe that the fact our method provides a reasonable accuracy on this benchmark is a very strong result.

% According to the initial estimation results shown in Table~\ref{tab:LINEMOD}, we can observe that our approach generalizes well to some novel objects (eggbox and glue) without any pose refinement. Furthermore, compared to the other state-of-the-art methods that trained separate network or individual branch for each object with real data(or a combination with synthetic data) following category-specific setting, our category-free approach trained only on synthetic data can obtain an comparable performance after the pose refinement provided by~\cite{li2018deepim}.

\begin{table}
	\centering
	\scriptsize \addtolength{\tabcolsep}{-4pt}
	\begin{tabular}{c l c c c c c c c c c c c c c |>{\columncolor{HL}}c}
		\toprule
		  \multicolumn{2}{l}{\textbf{LINEMOD}~\cite{hinterstoisser2012LINEMOD}} & ape & bvise & cam & can & cat & drill & duck & ebox\rlap*\, & glue\rlap*\, & holep & iron & lamp & phone & mean \\
		% & \#\,images & 1235 & 1214 & 1200 & 1195 & 1178 & 1187 & 1253 & 1252 & 1219 & 1236 & 1151 & 1226 & 1224 & 15770 \\ %18241 \\
% 		& \#\,models &
%         1 & 1 & 1 & 1 & 1 & 1 & 1 & 1 & 1 & 1 & 1 & 1 & 1 & 13 \\ 
		\midrule 
		& \multicolumn{15}{c}{\textbf{instance-specific networks/branches --- tested on supervised models \ (\ADD)*}} \\ % & \textbf{mean} \\ 
		\midrule
		
		% \multicolumn{16}{c}{\textbf{instance-specific on supervised instances in} \bm{$ADD-0.1d$}} \\ \midrule
		
		\multirow{5}{*}{w/o Ref.} & Brachmann~\cite{Brachmann2016UncertaintyDriven6P} & - & - & - & - & - & - & - & - & - & - & - & - & - & 32.3  \\
		& SSD-6D~\cite{SSD6D2017}$\ddag$ & 0 & 0.2 & 0.4 & 1.4 & 0.5 & 2.6 & 0 & 8.9 & 0 & 0.3 & 8.9 & 8.2 & 0.2 & 2.4 \\
		& BB8~\cite{rad2017bb8} & \textbf{27.9} & 62.0 & 40.1 & 48.1 & 45.2 & 58.6 & 32.8 & 40.0 & 27.0 & 42.4 & 67.0 & 39.9 & 35.2 & 43.6 \\
		& Tekin~\cite{tekin2018real} & 21.6 & \textbf{81.8} & 36.6 & 68.8 & 41.8 & 63.5 & 27.2 & 69.6 & 80.0 & 42.6 & \textbf{75.0} & \textbf{71.1} & 47.7 & 56.0 \\
		& PoseCNN~\cite{PoseCNN2018}$\dag$ & 27.8 & 68.9 & \textbf{47.5} & \textbf{71.4} & \textbf{56.7} & \textbf{65.4} & \textbf{42.8} & \textbf{98.3} & \textbf{95.2} & \textbf{50.9} & 65.6 & 70.3 & \textbf{54.6} & \textbf{62.7}  \\
		\midrule
		\multirow{4}{*}{w/ Ref.} & Brachmann~\cite{Brachmann2016UncertaintyDriven6P}& 33.2 & 64.8 & 38.4 & 62.9 & 42.7 & 61.9 & 30.2 & 49.9 & 31.2 & 52.8 & 80.0 & 67.0 & 38.1 & 50.2  \\
		& BB8~\cite{rad2017bb8} & 40.4 & 91.8 & 55.7 & 64.1 & 62.6 & 74.4 & 44.3 & 57.8 & 41.2 & \textbf{67.2} & 84.7 & 76.5 & 54.0 & 62.7 \\
	    & SSD-6D~\cite{SSD6D2017}$\ddag$ & 65 & 80 & 78 & 86 & 70 & 73 & 66 & \textbf{100} & \textbf{100} & 49 & 78 & 73 & 79 & 79.0 \\
		& PoseCNN~\cite{PoseCNN2018}$\dag$ + \cite{li2018deepim}$\dag$ & \textbf{76.9} & \textbf{97.4} & \textbf{93.5} & \textbf{96.6} & \textbf{82.1} & \textbf{95.0} & \textbf{77.7} & 97.0 & 99.4 & 52.7 & \textbf{98.3} & \textbf{97.5} & \textbf{87.8} & \textbf{88.6} \\
		
		\midrule & \multicolumn{15}{c}{\textbf{instance/category-agnostic network --- tested on novel models \ (\ADD)*}} \\ % & \textbf{mean} \\ 
		\midrule
		% \multicolumn{15}{c|}{\textbf{instance-agnostic on novel instances in} (ADD-0.1d)} & \textbf{mean} \\ \midrule
		w/o Ref. & Ours$\ddag$ & \bf 7.5 & \bf 25.1 & \bf 12.1 & \bf 11.3 & \bf 15.4 & \bf 18.6 & \bf 8.2 & \bf 100 & \bf 81.2 & \bf 18.5 & \bf 13.8 & \bf 6.5 & \bf 13.4 & \bf 25.5 \\ 
		\midrule
		w/ Ref. & Ours$\ddag$ + DeepIM~\cite{li2018deepim}$\dag$ & \bf 59.1 & \bf 63.8 & \bf 40.0 & \bf 50.8 & \bf 54.1 & \bf 75.3 & \bf 48.6 & \bf 100 & \bf 98.7 & \bf 49.8 & \bf 49.5 & \bf 55.3 & \bf 50.4 & \bf 61.2 \\
		%\midrule
		\bottomrule
	\multicolumn{16}{c}{[scenes: 13, artificially arranged | images: 13407 | objects: 13 | categ.: 13 | 3D models: 13, exact | align.: pixel]} \\
	\end{tabular}
    \vspace{1mm}
	\caption{Pose estimation on LINEMOD~\cite{hinterstoisser2012LINEMOD}. % 13 objects following common convention (15,770 images of 15 artificially arranged scenes, 15 objects in 15 categories, 15 ``exact'' 3D models, accurate alignments). 
	% Accuracy measured as ADD-0.1d (3D transformation distance < 10\% object dimension). 
	$\dag$~Training also on synthetic data.
	$\ddag$~Training only on synthetic data.
	*~\ADDS\ used for symmetric objects eggbox and glue. %*Results after the pose refinement method offered by DeepIM~\cite{li2018deepim}.}% All the previous methods are instance-specific while our approach is instance-agnostic.}
	% methods ranked by publication date, then rank by performance for ones with same publication date
	}
	\label{tab:LINEMOD}
\end{table}

\section{Conclusion}
\label{sec:conclu}

We have presented a new paradigm for deep pose estimation, taking the 3D object model as an input to the network. We demonstrated the benefits of this approach in terms of accuracy, and improved the state of the art on several standard pose estimation datasets. More importantly, we have shown that our approach holds the promise of a completely generic deep learning method for pose estimation, independent of the object category and training data, by showing encouraging results on the LINEMOD dataset without any specific training, and despite the domain gap between synthetic training data and real images for testing.
	
%\bibliography{main}

\begin{thebibliography}{54}
\providecommand{\natexlab}[1]{#1}
\providecommand{\url}[1]{\texttt{#1}}
\expandafter\ifx\csname urlstyle\endcsname\relax
  \providecommand{\doi}[1]{doi: #1}\else
  \providecommand{\doi}{doi: \begingroup \urlstyle{rm}\Url}\fi

\bibitem[Brachmann et~al.(2016)Brachmann, Michel, Krull, Yang, Gumhold, and
  Rother]{Brachmann2016UncertaintyDriven6P}
Eric Brachmann, Frank Michel, Alexander Krull, Michael~Ying Yang, Stefan
  Gumhold, and Carsten Rother.
\newblock Uncertainty-driven {6D} pose estimation of objects and scenes from a
  single {RGB} image.
\newblock In \emph{Conference on Computer Vision and Pattern Recognition
  (CVPR)}, 2016.

\bibitem[Chang et~al.(2015)Chang, Funkhouser, Guibas, Hanrahan, Huang, Li,
  Savarese, Savva, Song, Su, Xiao, Yi, and Yu]{shapenet2015}
Angel~X. Chang, Thomas Funkhouser, Leonidas Guibas, Pat Hanrahan, Qixing Huang,
  Zimo Li, Silvio Savarese, Manolis Savva, Shuran Song, Hao Su, Jianxiong Xiao,
  Li~Yi, and Fisher Yu.
\newblock {ShapeNet: An Information-Rich 3D Model Repository}.
\newblock Technical Report arXiv:1512.03012 [cs.GR], Stanford University --
  Princeton University -- Toyota Technological Institute at Chicago, 2015.

\bibitem[Deng et~al.(2009)Deng, Dong, Socher, Li, Li, and
  Fei-Fei]{imagenet_2009}
Jia Deng, Wei Dong, Richard Socher, Li-Jia Li, Kai Li, and Li~Fei-Fei.
\newblock {ImageNet: A Large-Scale Hierarchical Image Database}.
\newblock In \emph{Conference on Computer Vision and Pattern Recognition
  (CVPR)}, 2009.

\bibitem[Elhoseiny et~al.(2016)Elhoseiny, El-Gaaly, Bakry, and
  Elgammal]{Elhoseiny2016ACA}
Mohamed Elhoseiny, Tarek El-Gaaly, Amr Bakry, and Ahmed~M. Elgammal.
\newblock A comparative analysis and study of multiview {CNN} models for joint
  object categorization and pose estimation.
\newblock In \emph{International Conference on Machine Learning (ICML)}, 2016.

\bibitem[Engelmann et~al.(2017)Engelmann, Kontogianni, Hermans, and
  Leibe]{engelmann2017exploring}
Francis Engelmann, Theodora Kontogianni, Alexander Hermans, and Bastian Leibe.
\newblock Exploring spatial context for {3D} semantic segmentation of point
  clouds.
\newblock In \emph{International Conference on Computer Vision (ICCV)}, 2017.

\bibitem[Ferraz et~al.(2014)Ferraz, Binefa, and Moreno-Noguer]{ferraz2014very}
Luis Ferraz, Xavier Binefa, and Francesc Moreno-Noguer.
\newblock Very fast solution to the {PnP} problem with algebraic outlier
  rejection.
\newblock In \emph{Conference on Computer Vision and Pattern Recognition
  (CVPR)}, 2014.

\bibitem[Free3D()]{free3d}
Free3D.
\newblock Free3d.
\newblock \url{https://free3d.com}.

\bibitem[Geiger et~al.(2012)Geiger, Lenz, and Urtasun]{Geiger2012kitti}
Andreas Geiger, Philip Lenz, and Raquel Urtasun.
\newblock Are we ready for autonomous driving? the {KITTI} vision benchmark
  suite.
\newblock In \emph{Conference on Computer Vision and Pattern Recognition
  (CVPR)}, 2012.

\bibitem[Georgakis et~al.(2018)Georgakis, Karanam, Wu, and
  Kosecka]{MatchingRI2018}
Georgios Georgakis, Srikrishna Karanam, Ziyan Wu, and Jana Kosecka.
\newblock Matching {RGB} images to {CAD} models for object pose estimation.
\newblock \emph{CoRR}, abs/1811.07249, 2018.

\bibitem[Grabner et~al.(2018)Grabner, Roth, and Lepetit]{3Dpose3Dmodel2018}
Alexander Grabner, Peter~M. Roth, and Vincent Lepetit.
\newblock {3D} pose estimation and {3D} model retrieval for objects in the
  wild.
\newblock In \emph{Conference on Computer Vision and Pattern Recognition
  (CVPR)}, 2018.

\bibitem[Groueix et~al.(2018)Groueix, Fisher, Kim, Russell, and
  Aubry]{atlasnet2018}
Thibault Groueix, Matthew Fisher, Vladimir~G. Kim, Bryan Russell, and Mathieu
  Aubry.
\newblock {AtlasNet}: A papier-m\^ach\'e approach to learning {3D} surface
  generation.
\newblock In \emph{Conference on Computer Vision and Pattern Recognition
  (CVPR)}, 2018.

\bibitem[G\"uler et~al.(2017)G\"uler, Trigeorgis, Antonakos, Snape, Zafeiriou,
  and Kokkinos]{Guler2017}
R{\i}za~Alp G\"uler, George Trigeorgis, Epameinondas Antonakos, Patrick Snape,
  Stefanos Zafeiriou, and Iasonas Kokkinos.
\newblock Densereg: Fully convolutional dense shape regression in-the-wild.
\newblock In \emph{Conference on Computer Vision and Pattern Recognition
  (CVPR)}, 2017.

\bibitem[He et~al.(2016)He, Zhang, Ren, and Sun]{He2016DeepRL}
Kaiming He, Xiangyu Zhang, Shaoqing Ren, and Jian Sun.
\newblock Deep residual learning for image recognition.
\newblock In \emph{Conference on Computer Vision and Pattern Recognition
  (CVPR)}, 2016.

\bibitem[Hinterstoisser et~al.(2012{\natexlab{a}})Hinterstoisser, Cagniart,
  Ilic, Sturm, Navab, Fua, and Lepetit]{hinterstoisser2012gradient}
Stefan Hinterstoisser, Cedric Cagniart, Slobodan Ilic, Peter Sturm, Nassir
  Navab, Pascal Fua, and Vincent Lepetit.
\newblock Gradient response maps for real-time detection of textureless
  objects.
\newblock \emph{IEEE Transactions on Pattern Analysis and Machine Intelligence
  (PAMI)}, 2012{\natexlab{a}}.

\bibitem[Hinterstoisser et~al.(2012{\natexlab{b}})Hinterstoisser, Lepetit,
  Ilic, Holzer, Bradski, Konolige, and Navab]{hinterstoisser2012LINEMOD}
Stefan Hinterstoisser, Vincent Lepetit, Slobodan Ilic, Stefan Holzer, Gary
  Bradski, Kurt Konolige, and Nassir Navab.
\newblock Model based training, detection and pose estimation of texture-less
  3d objects in heavily cluttered scenes.
\newblock In \emph{Asian Conference on Computer Vision (ACCV)},
  2012{\natexlab{b}}.

\bibitem[Huber(1992)]{huberloss}
Peter~J Huber.
\newblock Robust estimation of a location parameter.
\newblock In \emph{Breakthroughs in statistics}. Springer New York, 1992.

\bibitem[Kehl et~al.(2017)Kehl, Manhardt, Tombari, Ilic, and Navab]{SSD6D2017}
Wadim Kehl, Fabian Manhardt, Federico Tombari, Slobodan Ilic, and Nassir Navab.
\newblock {SSD-6D}: Making {RGB}-based {3D} detection and {6D} pose estimation
  great again.
\newblock In \emph{International Conference on Computer Vision (ICCV)}, 2017.

\bibitem[Kendall and Cipolla(2017)]{kendall2017geometricpose}
Alex Kendall and Roberto Cipolla.
\newblock Geometric loss functions for camera pose regression with deep
  learning.
\newblock In \emph{Conference on Computer Vision and Pattern Recognition
  (CVPR)}, 2017.

\bibitem[Kendall et~al.(2015)Kendall, Grimes, and Cipolla]{posenet2015}
Alex Kendall, Matthew~Koichi Grimes, and Roberto Cipolla.
\newblock {PoseNet}: A convolutional network for real-time 6-{DOF} camera
  relocalization.
\newblock In \emph{International Conference on Computer Vision (ICCV)}, 2015.

\bibitem[Kingma and Ba(2014)]{Kingma2015AdamAM}
Diederik Kingma and Jimmy Ba.
\newblock Adam: A method for stochastic optimization.
\newblock In \emph{International Conference on Learning Representations
  (ICLR)}, 2014.

\bibitem[Lepetit et~al.(2009)Lepetit, Moreno-Noguer, and Fua]{lepetit2009epnp}
Vincent Lepetit, Francesc Moreno-Noguer, and Pascal Fua.
\newblock {EPnP}: An accurate {O}(n) solution to the {PnP} problem.
\newblock \emph{International Journal of Computer Vision (IJCV)}, 2009.

\bibitem[Li et~al.(2018{\natexlab{a}})Li, Bai, and Hager]{UnifiedMVMC2018}
Chi Li, Jin Bai, and Gregory~D. Hager.
\newblock A unified framework for multi-view multi-class object pose
  estimation.
\newblock In \emph{European Conference on Computer Vision (ECCV)},
  2018{\natexlab{a}}.

\bibitem[Li et~al.(2011)Li, Wang, Yin, and Wang]{li2011deformable}
Dengwang Li, Hongjun Wang, Yong Yin, and Xiuying Wang.
\newblock Deformable registration using edge-preserving scale space for
  adaptive image-guided radiation therapy.
\newblock \emph{Journal of Applied Clinical Medical Physics (JACMP)}, 2011.

\bibitem[Li et~al.(2012)Li, Xu, and Xie]{li2012robust}
Shiqi Li, Chi Xu, and Ming Xie.
\newblock A robust {O}(n) solution to the perspective-n-point problem.
\newblock \emph{IEEE Transactions on Pattern Analysis and Machine Intelligence
  (PAMI)}, 2012.

\bibitem[Li et~al.(2018{\natexlab{b}})Li, Wang, Ji, Xiang, and
  Fox]{li2018deepim}
Yi~Li, Gu~Wang, Xiangyang Ji, Yu~Xiang, and Dieter Fox.
\newblock {DeepIM}: Deep iterative matching for {6D} pose estimation.
\newblock In \emph{European Conference on Computer Vision (ECCV)},
  2018{\natexlab{b}}.

\bibitem[Lowe(1991)]{lowe1991fitting}
David~G. Lowe.
\newblock Fitting parameterized three-dimensional models to images.
\newblock \emph{IEEE Transactions on Pattern Analysis and Machine Intelligence
  (PAMI)}, 1991.

\bibitem[Lowe(2004)]{lowe2004distinctive}
David~G Lowe.
\newblock Distinctive image features from scale-invariant keypoints.
\newblock \emph{International Journal of Computer Vision (IJCV)}, 2004.

\bibitem[Mahendran et~al.(2018)Mahendran, Ali, and Vidal]{Mahendran2018AMC}
Siddharth Mahendran, Haider Ali, and Ren{\'e} Vidal.
\newblock A mixed classification-regression framework for {3D} pose estimation
  from {2D} images.
\newblock In \emph{British Machine Vision Conference (BMVC)}, 2018.

\bibitem[Manhardt et~al.(2018)Manhardt, Kehl, Navab, and
  Tombari]{manhardt2018modelrefine}
Fabian Manhardt, Wadim Kehl, Nassir Navab, and Federico Tombari.
\newblock Deep model-based {6D} pose refinement in {RGB}.
\newblock In \emph{European Conference on Computer Vision (ECCV)}, 2018.

\bibitem[Massa et~al.(2016)Massa, Marlet, and Aubry]{MultiTask2016}
Francisco Massa, Renaud Marlet, and Mathieu Aubry.
\newblock Crafting a multi-task cnn for viewpoint estimation.
\newblock In \emph{British Machine Vision Conference (BMVC)}, 2016.

\bibitem[Mousavian et~al.(2017)Mousavian, Anguelov, Flynn, and
  Kosecka]{3DBboxMousavian2017}
Arsalan Mousavian, Dragomir Anguelov, John Flynn, and Jana Kosecka.
\newblock {3D} bounding box estimation using deep learning and geometry.
\newblock In \emph{Conference on Computer Vision and Pattern Recognition
  (CVPR)}, 2017.

\bibitem[Oberweger et~al.(2018)Oberweger, Rad, and
  Lepetit]{oberweger2018heatmapspose}
Markus Oberweger, Mahdi Rad, and Vincent Lepetit.
\newblock Making deep heatmaps robust to partial occlusions for {3D} object
  pose estimation.
\newblock In \emph{European Conference on Computer Vision (ECCV)}, 2018.

\bibitem[Osadchy et~al.(2007)Osadchy, Cun, and Miller]{Osadchy2007}
Margarita Osadchy, Yann~Le Cun, and Matthew~L. Miller.
\newblock Synergistic face detection and pose estimation with energy-based
  models.
\newblock \emph{Journal of Machine Learning Research (JMLR)}, 2007.

\bibitem[Pavlakos et~al.(2017)Pavlakos, Zhou, Chan, Derpanis, and
  Daniilidis]{pavlakos20176}
Georgios Pavlakos, Xiaowei Zhou, Aaron Chan, Konstantinos~G Derpanis, and
  Kostas Daniilidis.
\newblock 6-dof object pose from semantic keypoints.
\newblock In \emph{International Conference on Robotics and Automation (ICRA)},
  2017.

\bibitem[Penedones et~al.(2012)Penedones, Collobert, Fleuret, and
  Grangier]{Penedones2012ImprovingOC}
Hugo Penedones, Ronan Collobert, Fran\c{c}ois Fleuret, and David Grangier.
\newblock Improving object classification using pose information.
\newblock Technical report, Idiap Research Institute, 2012.

\bibitem[Qi et~al.(2018)Qi, Liu, Wu, Su, and Guibas]{qi2018frustumpointnet}
Charles~R Qi, Wei Liu, Chenxia Wu, Hao Su, and Leonidas~J Guibas.
\newblock Frustum pointnets for {3D} object detection from {RGB-D} data.
\newblock In \emph{Conference on Computer Vision and Pattern Recognition
  (CVPR)}, 2018.

\bibitem[Qi et~al.(2017{\natexlab{a}})Qi, Su, Mo, and Guibas]{Qi2017PointNetDL}
Charles~Ruizhongtai Qi, Hao Su, Kaichun Mo, and Leonidas~J. Guibas.
\newblock {PointNet}: Deep learning on point sets for {3D} classification and
  segmentation.
\newblock In \emph{Conference on Computer Vision and Pattern Recognition
  (CVPR)}, 2017{\natexlab{a}}.

\bibitem[Qi et~al.(2017{\natexlab{b}})Qi, Yi, Su, and Guibas]{Qi2017PointNetDH}
Charles~Ruizhongtai Qi, Li~Yi, Hao Su, and Leonidas~J. Guibas.
\newblock {PointNet++}: Deep hierarchical feature learning on point sets in a
  metric space.
\newblock In \emph{Conference on Neural Information Processing Systems (NIPS)},
  2017{\natexlab{b}}.

\bibitem[Rad and Lepetit(2017)]{rad2017bb8}
Mahdi Rad and Vincent Lepetit.
\newblock {BB8}: a scalable, accurate, robust to partial occlusion method for
  predicting the {3D} poses of challenging objects without using depth.
\newblock In \emph{International Conference on Computer Vision (ICCV)}, 2017.

\bibitem[Su et~al.(2015{\natexlab{a}})Su, Maji, Kalogerakis, and
  Learned{-}Miller]{mvcnn2015}
Hang Su, Subhransu Maji, Evangelos Kalogerakis, and Erik~G. Learned{-}Miller.
\newblock Multi-view convolutional neural networks for 3d shape recognition.
\newblock In \emph{International Conference on Computer Vision (ICCV)},
  2015{\natexlab{a}}.

\bibitem[Su et~al.(2015{\natexlab{b}})Su, Qi, Li, and Guibas]{RenderForCNN2015}
Hao Su, Charles~R. Qi, Yangyan Li, and Leonidas~J. Guibas.
\newblock Render for {CNN}: Viewpoint estimation in images using {CNNs} trained
  with rendered {3D} model views.
\newblock In \emph{International Conference on Computer Vision (ICCV)},
  2015{\natexlab{b}}.

\bibitem[Sun et~al.(2018)Sun, Wu, Zhang, Zhang, Zhang, Xue, Tenenbaum, and
  Freeman]{pix3d2018}
Xingyuan Sun, Jiajun Wu, Xiuming Zhang, Zhoutong Zhang, Chengkai Zhang, Tianfan
  Xue, Joshua~B Tenenbaum, and William~T Freeman.
\newblock {Pix3D}: Dataset and methods for single-image 3d shape modeling.
\newblock In \emph{Conference on Computer Vision and Pattern Recognition
  (CVPR)}, 2018.

\bibitem[Tekin et~al.(2018)Tekin, Sinha, and Fua]{tekin2018real}
Bugra Tekin, Sudipta~N Sinha, and Pascal Fua.
\newblock Real-time seamless single shot {6D} object pose prediction.
\newblock In \emph{Conference on Computer Vision and Pattern Recognition
  (CVPR)}, 2018.

\bibitem[Tjaden et~al.(2017)Tjaden, Schwanecke, and
  Sch{\"o}mer]{RealTimeMP2017}
Henning Tjaden, Ulrich Schwanecke, and Elmar Sch{\"o}mer.
\newblock Real-time monocular pose estimation of {3D} objects using temporally
  consistent local color histograms.
\newblock In \emph{International Conference on Computer Vision (ICCV)}, 2017.

\bibitem[Tola et~al.(2010)Tola, Lepetit, and Fua]{tola2010daisy}
Engin Tola, Vincent Lepetit, and Pascal Fua.
\newblock Daisy: An efficient dense descriptor applied to wide-baseline stereo.
\newblock \emph{IEEE Transactions on Pattern Analysis and Machine Intelligence
  (PAMI)}, 2010.

\bibitem[Tulsiani and Malik(2015)]{ViewpointsKeypoints2015}
Shubham Tulsiani and Jitendra Malik.
\newblock Viewpoints and keypoints.
\newblock In \emph{Conference on Computer Vision and Pattern Recognition
  (CVPR)}, 2015.

\bibitem[Wang et~al.(2018)Wang, Sun, Liu, Sarma, Bronstein, and
  Solomon]{wang2018dynamichgraphcnn}
Yue Wang, Yongbin Sun, Ziwei Liu, Sanjay~E Sarma, Michael~M Bronstein, and
  Justin~M Solomon.
\newblock Dynamic graph {CNN} for learning on point clouds.
\newblock \emph{arXiv preprint arXiv:1801.07829}, 2018.

\bibitem[Xiang et~al.(2014)Xiang, Mottaghi, and Savarese]{pascal3d2014}
Yu~Xiang, Roozbeh Mottaghi, and Silvio Savarese.
\newblock Beyond {PASCAL}: A benchmark for {3D} object detection in the wild.
\newblock In \emph{Winter Conference on Applications of Computer Vision
  (WACV)}, 2014.

\bibitem[Xiang et~al.(2016)Xiang, Kim, Chen, Ji, Choy, Su, Mottaghi, Guibas,
  and Savarese]{object3d2016}
Yu~Xiang, Wonhui Kim, Wei Chen, Jingwei Ji, Christopher Choy, Hao Su, Roozbeh
  Mottaghi, Leonidas Guibas, and Silvio Savarese.
\newblock {ObjectNet3D}: A large scale database for {3D} object recognition.
\newblock In \emph{European Conference Computer Vision (ECCV)}, 2016.

\bibitem[Xiang et~al.(2018)Xiang, Schmidt, Narayanan, and Fox]{PoseCNN2018}
Yu~Xiang, Tanner Schmidt, Venkatraman Narayanan, and Dieter Fox.
\newblock {PoseCNN}: A convolutional neural network for {6D} object pose
  estimation in cluttered scenes.
\newblock In \emph{Robotics: Science and Systems (RSS)}, 2018.

\bibitem[{Xiao} et~al.(2010){Xiao}, {Hays}, {Ehinger}, {Oliva}, and
  {Torralba}]{SUN2010}
J.~{Xiao}, J.~{Hays}, K.~A. {Ehinger}, A.~{Oliva}, and A.~{Torralba}.
\newblock {SUN} database: Large-scale scene recognition from abbey to zoo.
\newblock In \emph{Conference on Computer Vision and Pattern Recognition
  (CVPR)}, 2010.

\bibitem[Xu et~al.(2018)Xu, Anguelov, and Jain]{xu2018pointfusion}
Danfei Xu, Dragomir Anguelov, and Ashesh Jain.
\newblock Pointfusion: Deep sensor fusion for {3D} bounding box estimation.
\newblock In \emph{Conference on Computer Vision and Pattern Recognition
  (CVPR)}, 2018.

\bibitem[Zheng et~al.(2013)Zheng, Kuang, Sugimoto, Astrom, and
  Okutomi]{zheng2013revisiting}
Yinqiang Zheng, Yubin Kuang, Shigeki Sugimoto, Kalle Astrom, and Masatoshi
  Okutomi.
\newblock Revisiting the {PnP} problem: A fast, general and optimal solution.
\newblock In \emph{International Conference on Computer Vision (ICCV)}, 2013.

\bibitem[Zhou et~al.(2018)Zhou, Karpur, Luo, and Huang]{starmap2018}
Xingyi Zhou, Arjun Karpur, Linjie Luo, and Qixing Huang.
\newblock Starmap for category-agnostic keypoint and viewpoint estimation.
\newblock In \emph{European Conference on Computer Vision (ECCV)}, 2018.

\end{thebibliography}

\newpage
\section{Supplementary Material}

\subsection{Datasets}
\paragraph{Pascal3D+}~\cite{pascal3d2014} provides images with 3D annotations for 12 object categories. The images are selected from the training and validation set of PASCAL VOC 2012~\cite{pascal-voc-2012} and ImageNet \cite{imagenet_2009}, with 2k to 4k images in the wild per category. An approximate 3D CAD model is provided for each object as well as its 3D orientation in the image. Following the protocol of \cite{ViewpointsKeypoints2015,3DBboxMousavian2017,3Dpose3Dmodel2018}% and others
, we use the ImageNet-trainval and Pascal-train images as training data, and the 2,113 non-occluded and non-truncated objects of the Pascal-val images as testing data. As in~\cite{ViewpointsKeypoints2015}, we use the metric $\Accpisix$, which measures the percentage of test samples having a pose prediction error smaller than $\frac{\pi}{6}$: $\Delta(R_\pred, R_\gt) = \relax{\| \log (R_\pred^{T} R_\gt)\|_{\mathcal{F}}}/{\sqrt{2}} < \frac{\pi}{6}$.

\paragraph{ObjectNet3D}~\cite{object3d2016} is a large-scale 3D dataset similar to Pascal3D+ but with 100 categories, which provide a wider variety of shapes. To verify the generalization power of our method for unknown categories, we follow the protocol of StarMap~\cite{starmap2018}: we evenly hold out 20 categories (every 5 categories sorted in the alphabetical order) from the training data and only used them for testing.
%bed, bookcase (bcase), calculator (calc), cellphone (cphone), computer (comp), door, cabinet (cabi), guitar (guit), iron, knife, microwave (micro), pen, pot, rifle, shoe, slipper, stove, toilet, tub and wheelchair (wchair). Among the 50k training samples provided in ObjectNet3D, \cite{starmap2018} only used a 20k subset of them which contain the keypoint annotations since their method relies on the 2D-3D keypoint correspondence for network training.
For a fair comparison, we actually use the same subset of training data as in~\cite{object3d2016} (also containing keypoint annotations) and evaluate on the non-occluded and non-truncated images of the 20 categories, using the same $\Accpisix$ metric.
%In total, we collect 19k images for training and 4k images for novel categories. 

\paragraph{Pix3D}~\cite{pix3d2018} is a recent dataset containing 5,711 non-occluded and non-truncated images of 395 CAD shapes among 9 categories. It mainly features furniture, with a strong bias towards chairs. But contrary to Pascal3D+ and ObjectNet3D, that only feature approximate models and rough alignments, Pix3D provides exact models and pixel-level accurate poses. Similar to the training paradigm of~\cite{RenderForCNN2015,pix3d2018}, we train on ShapeNetCore~\cite{shapenet2015} with input images made of rendered views on random SUN397 backgrounds~\cite{SUN2010} using random texture maps included in ShapeNetCore, and test on Pix3D real images and shapes.

\paragraph{ShapeNetCore} is a subset of ShapeNet~\cite{shapenet2015} containing 51k single clean 3D models, covering 55 common object categories of man-made artifacts. We exclude the categories containing mostly objects with rotational symmetry or small and narrow objects, which results in 30 remaining categories: \emph{airplane, bag, bathtub, bed, birdhouse, bookshelf, bus, cabinet, camera, car, chair, clock, dishwasher, display, faucet, lamp, laptop, speaker, mailbox, microwave, motorcycle, piano, pistol, printer, rifle, sofa, table, train, watercraft} and \emph{washer}. We randomly choose 200 models from each category and use Blender to render each model under 20 random views with various textures included in ShapeNetCore.

\paragraph{LINEMOD}~\cite{hinterstoisser2012LINEMOD} has become a standard benchmark for 6D pose estimation of textureless objects in cluttered scenes. It consists of 15 sequences featuring one object instance for each sequence to detect with ground truth 6D pose and object class. As other authors, we left out categories bowl and cup, that have a rotational symmetry, and consider only 13 classes. 
The common evaluation measure with LINEMOD is the ADD-0.1d metric~\cite{hinterstoisser2012LINEMOD}: a pose is considered correct if the average of the 3D distances between transformed object vertices by the ground truth transformation and the ones by estimated transformation is less than 10\% of the object's diameter. For the objects with ambiguous poses due to symmetries, \cite{hinterstoisser2012LINEMOD} replaces this measure by ADD-S which is specially tailored for symmetric objects. We choose ADD-0.1d and ADD-S-0.1d as our evaluation metrics.

\subsection{Evaluation Metrics}
For results on LINEMOD, the ADD~\cite{hinterstoisser2012LINEMOD} metric is used to compute the averaged distance between points transformed using the estimated pose and the ground truth pose:
\begin{equation}
\textbf{ADD} = \frac{1}{m} \sum_{\mathbf{x} \in \mathcal{M}} || (\mathbf{Rx} + \mathbf{t}) - (\mathbf{\hat{R}x} + \mathbf{\hat{t}}) ||
\end{equation}
where $m$ is the number of points on the 3D object model, $\mathcal{M}$ is the set of all 3D points of this model, $\mathbf{p} = [\mathbf{R} | \mathbf{t}]$ is the ground truth pose and $\mathbf{\hat{p}} = [\mathbf{\hat{R}} | \mathbf{\hat{t}}]$ is the estimated pose. Following~\cite{Brachmann2016UncertaintyDriven6P}, we compute the model diameter $d$ as the maximum distance between all pairs of points from the model. With this metric, a pose estimation is considered to be correct if the computed averaged distance is within 10\% of the model diameter $d$.

For the objects with ambiguous poses due to symmetries, \cite{hinterstoisser2012LINEMOD} replaces this measure by ADD-S, which uses the closet point distance in computing the average distance for 6D pose evaluation as in:
\begin{equation}
\textbf{ADD-S} = \frac{1}{m} \sum_{\mathbf{x}_1 \in \mathcal{M}} \min_{\mathbf{x}_2 \in \mathcal{M}} || (\mathbf{Rx}_1 + \mathbf{t}) - (\mathbf{\hat{R}x}_2 + \mathbf{\hat{t}}) ||
\end{equation}

%-------------------------------------------------------------------------
\subsection{Ablation and parameter study}

\paragraph{Ablation and parameter study on the number of rendered images.} Table~\ref{tab:ablation_view} shows the experimental results of pose estimation on 20 novel categories of ObjectNet3D for different numbers and layouts of rendered images. The viewpoints are sampled evenly at $N_{\azi}$ azimuths and elevated at $N_{\ele}$ different elevations. $N_{\ele}=1, 2, 3$ represents respectively elevations at $(30\degree)$, $(0\degree, 30\degree)$, $(0\degree, 30\degree, 60\degree)$. The $\Accpisix$ metric measures the percentage of testing samples with a angular error smaller than $\frac{\pi}{6}$ and \MedErr\ is the median angular error (\degree) over all testing samples.

The table shows that using shape information encoded from rendered images (when $N_{\azi} \times N_{\ele} > 0$) can indeed help pose estimation on novel categories, i.e., that are not included in the training data. In the first column (0 rendered images) we show the performance of our baseline without using the 3D shape of the object, compared to this result, the network trained with only one rendered image has a clearly boosted accuracy.

The table also shows that more rendered images in the network input does not necessarily mean a better performance. In the table, the network trained with 12 rendered images elevated at 0\degree and 30\degree gives the best result. This may be because the ObjectNet3D dataset is highly biased towards low elevations on the hemisphere, which can be well represented without using the rendered image captured at high elevation such as 60\degree.

\begin{table*}
	\centering
	\scriptsize 
	%\addtolength{\tabcolsep}{-3pt}
	\begin{tabular}{c|c|c| c c c| c c c| c c c}
		\toprule
		%Number & 0 & 1 & \multicolumn{3}{c|}{6} & \multicolumn{3}{c|}{12} & \multicolumn{3}{c}{18} \\
		$N_{\azi}\times N_{\ele}$ & 0 & 1$\times$1 & 6$\times$1 & 3$\times$2 & 2$\times$3 & 12$\times$1 & 6$\times$2 & 4$\times$3 & 18$\times$1 & 9$\times$2 & 6$\times$3 \\ \midrule
		$\Accpisix \uparrow$ & 50 & 56 & 59 & 60 & 58 & 59 & \textbf{62} & 58 & 58 & 60 & 59 \\[1mm]
		$MedErr \downarrow$ & 50 & 45 & 44 & 44 & 51 & 46 & \textbf{40} & 46 & 51 & 43 & 45 \\
		\bottomrule	
	\end{tabular}
	\vspace{2mm}
	\caption{Ablation and parameter study on ObjectNet3D of the number and layout of rendering images at the input of the network when using multiple views to represent shape. Performance depending on the number of azimuthal and elevation samples. 
		%The table shows the results if we change the sample of Azimuth and Elevation: $N_{\azi} \times N_{\ele}$.
	}
	\label{tab:ablation_view}
\end{table*}

\paragraph{Parameter study on the azimuthal randomization strategy.} Table~\ref{tab:ablation_random} summarizes the parameter study on the range of azimuthal jittering
applied to input shapes during network training. The poor results obtained for $[{-}0\degree,0\degree]$ and $[{-}180\degree,180\degree]$ are due the objects with symmetries, typically at 90\degree or 180\degree.
% Comparing the first column and the last column shows that the complete azimuth changes within $[{-}180\degree,180\degree]$ result in worse result, which is caused by the objects with symmetries, typically at 90\degree or 180\degree.

\begin{table}
	\centering
	\scriptsize
	\begin{tabular}{c|c |c |c |c}
		\toprule
		Randomization Range & $[{-}0\degree,0\degree]$ & $[{-}45\degree,45\degree]$ & $[{-}90\degree,90\degree]$ & $[{-}180\degree,180\degree]$\\
		\midrule
		$\Accpisix \uparrow$ & 56 & \textbf{62} & 60 & 55 \\[1mm]
		$MedErr \downarrow$ & 47 & \textbf{40} & 43 & 52 \\
		\bottomrule
	\end{tabular}
	\vspace{2mm}
	\caption{Parameter study of azimuthal randomization used as a specific data augmentation of our approach. Performance depending on the range of azimuthal variation during training.}
	\label{tab:ablation_random}
\end{table}

%-------------------------------------------------------------------------
\subsection{Qualitative Results on LINEMOD}
\begin{figure}
	\centering
	\begin{tabular}{cc}
		\bmvaHangBox{\includegraphics[width=.45\linewidth]{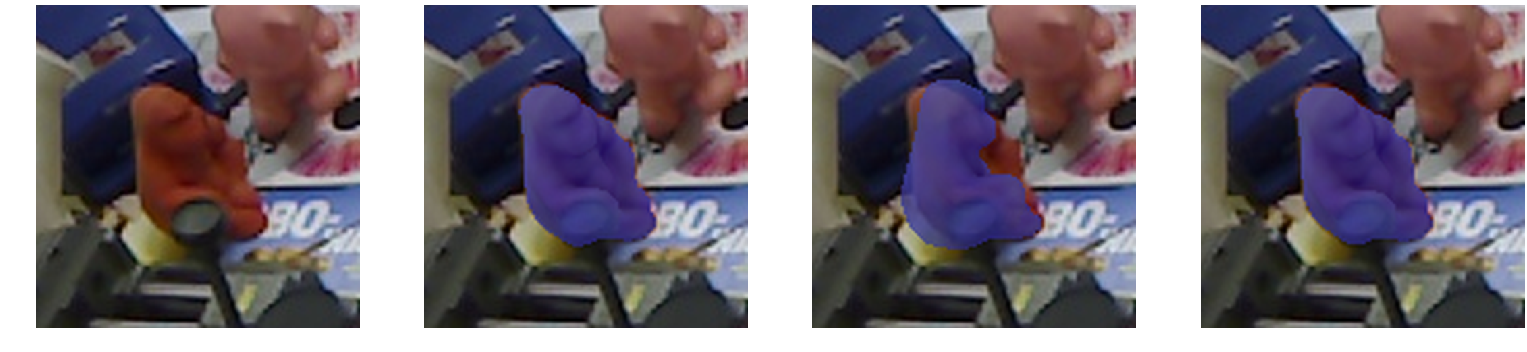}}&
		\bmvaHangBox{\includegraphics[width=.45\linewidth]{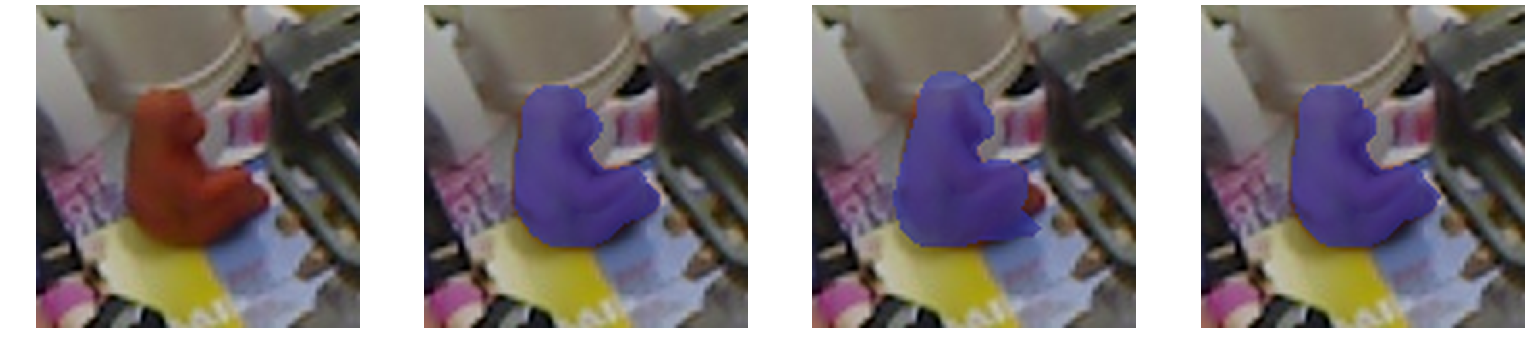}}\\
		\bmvaHangBox{\includegraphics[width=.45\linewidth]{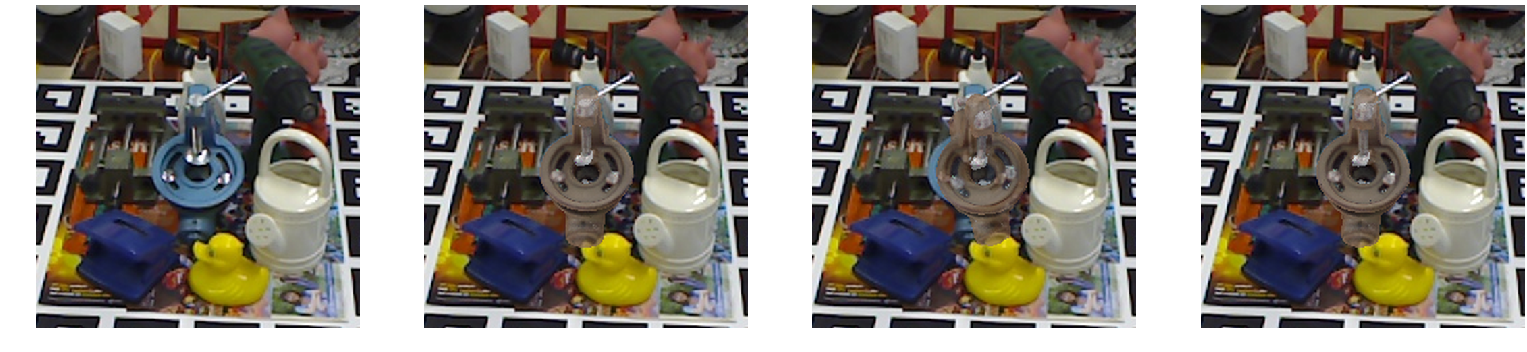}}&
		\bmvaHangBox{\includegraphics[width=.45\linewidth]{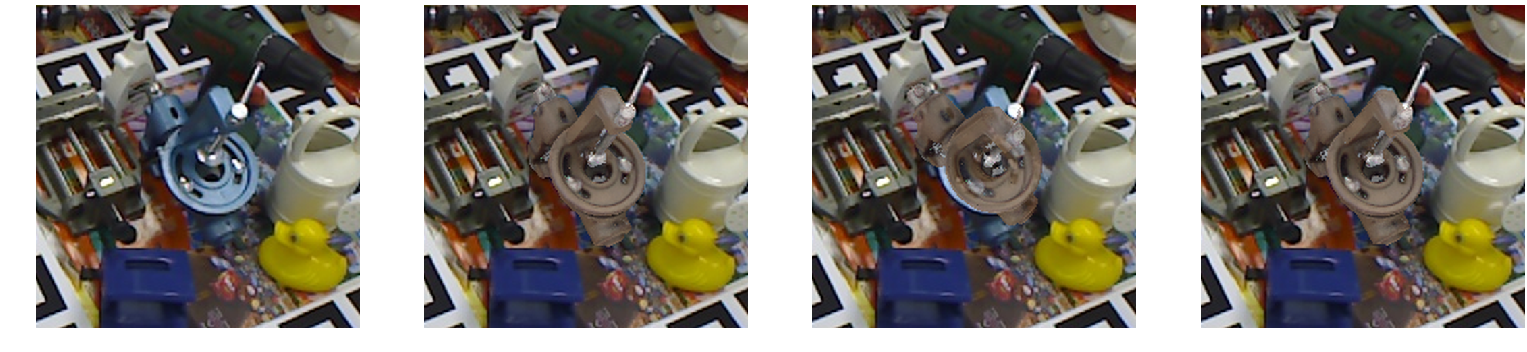}}\\
		\bmvaHangBox{\includegraphics[width=.45\linewidth]{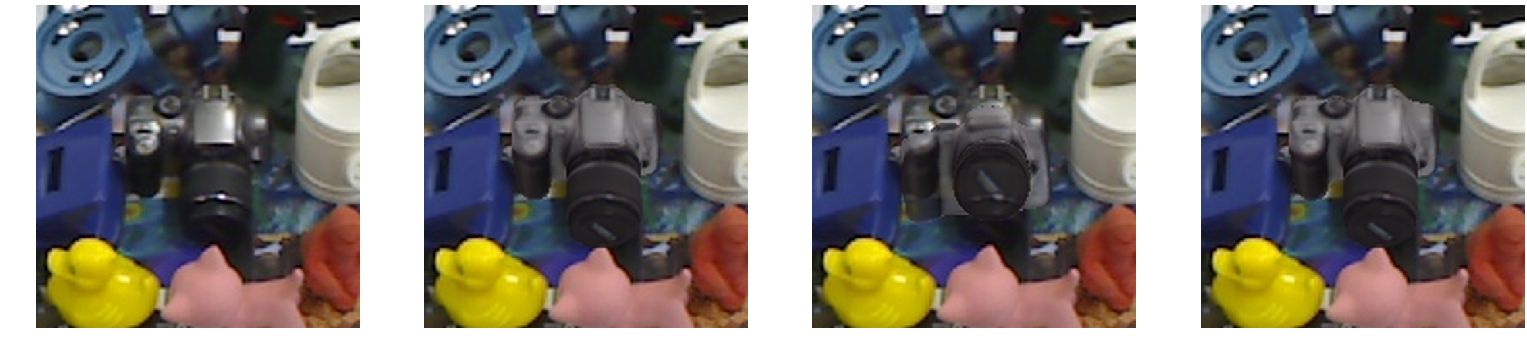}}&
		\bmvaHangBox{\includegraphics[width=.45\linewidth]{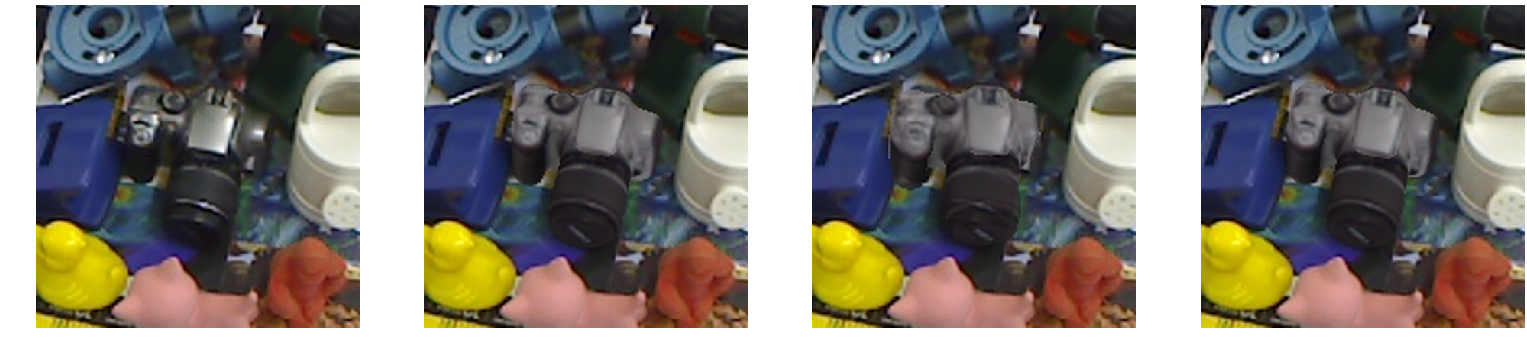}}\\
		\bmvaHangBox{\includegraphics[width=.45\linewidth]{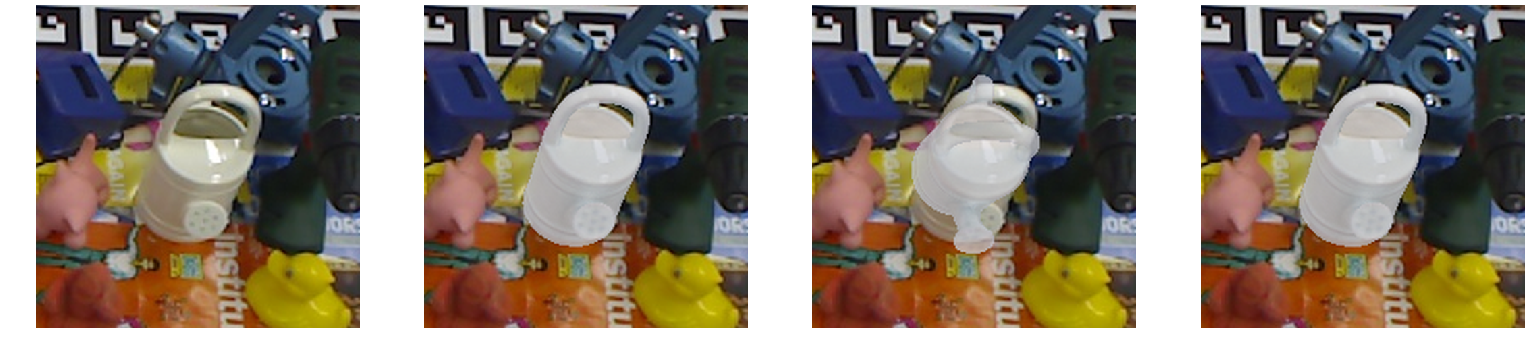}}&
		\bmvaHangBox{\includegraphics[width=.45\linewidth]{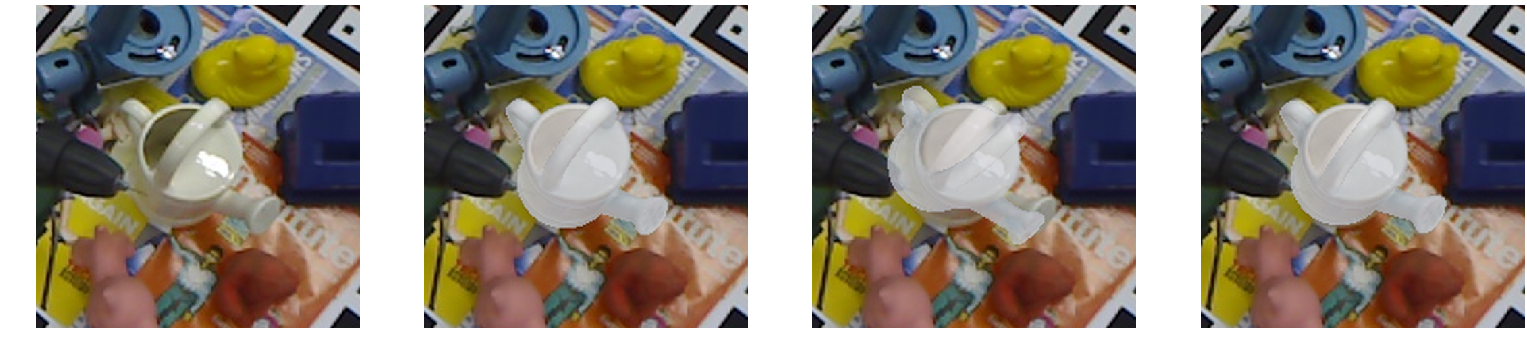}}\\
		\bmvaHangBox{\includegraphics[width=.45\linewidth]{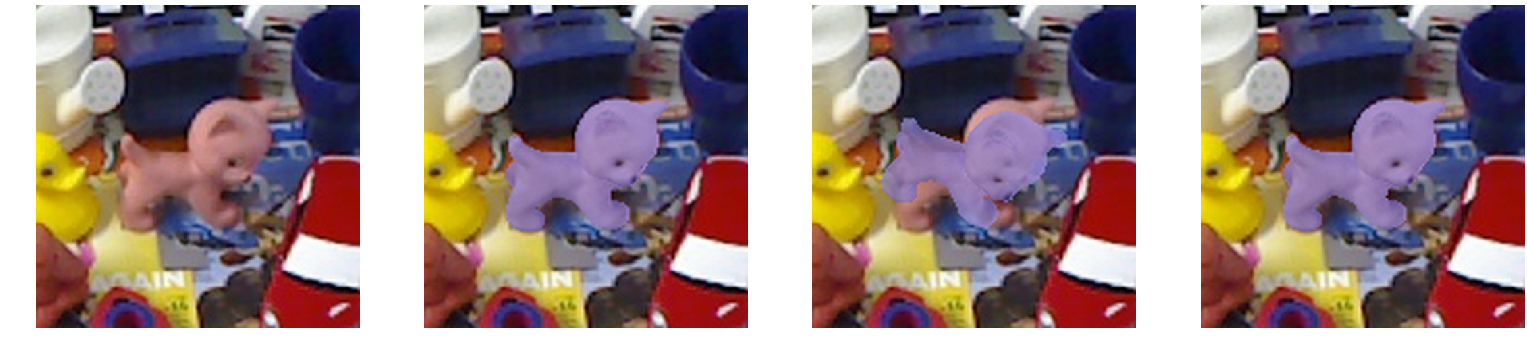}}&
		\bmvaHangBox{\includegraphics[width=.45\linewidth]{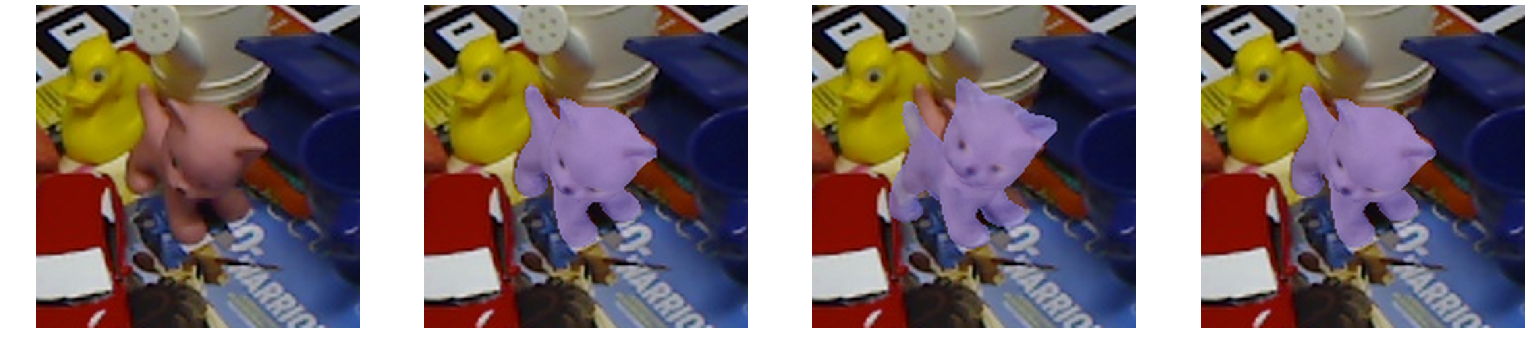}}\\
		\bmvaHangBox{\includegraphics[width=.45\linewidth]{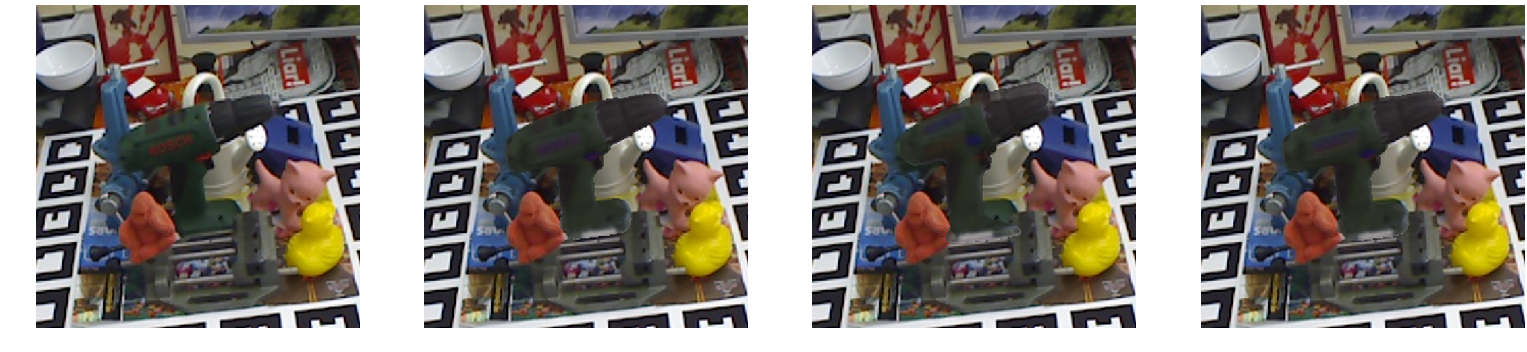}}&
		\bmvaHangBox{\includegraphics[width=.45\linewidth]{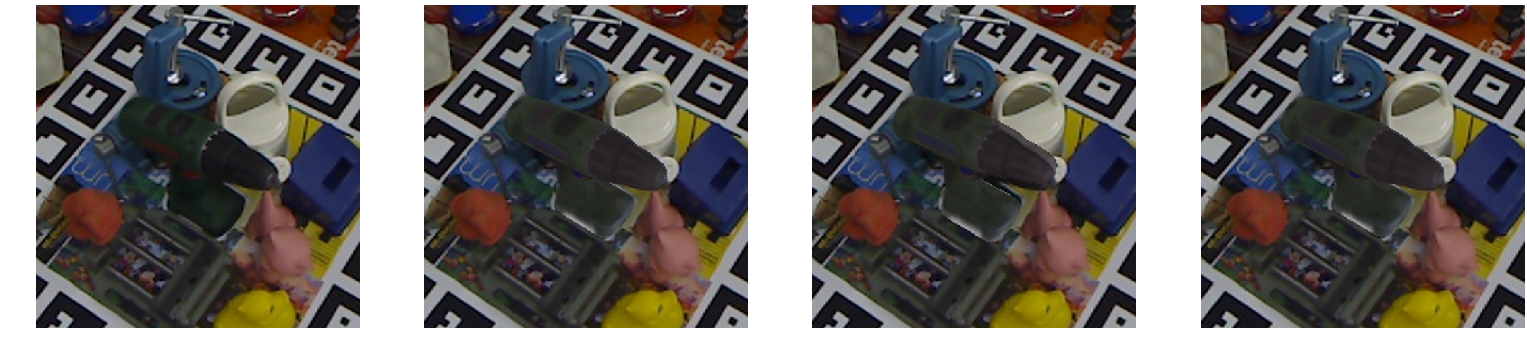}}\\
		\bmvaHangBox{\includegraphics[width=.45\linewidth]{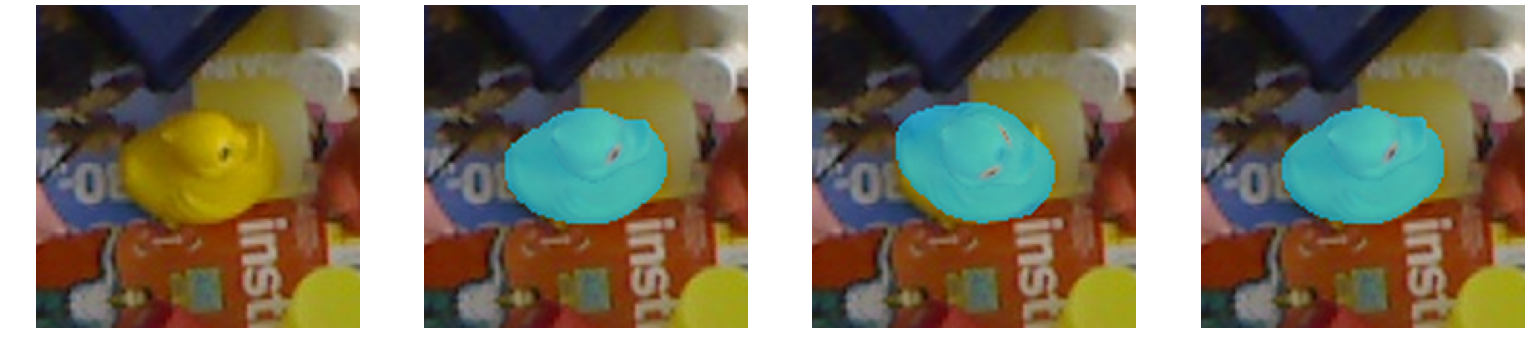}}&
		\bmvaHangBox{\includegraphics[width=.45\linewidth]{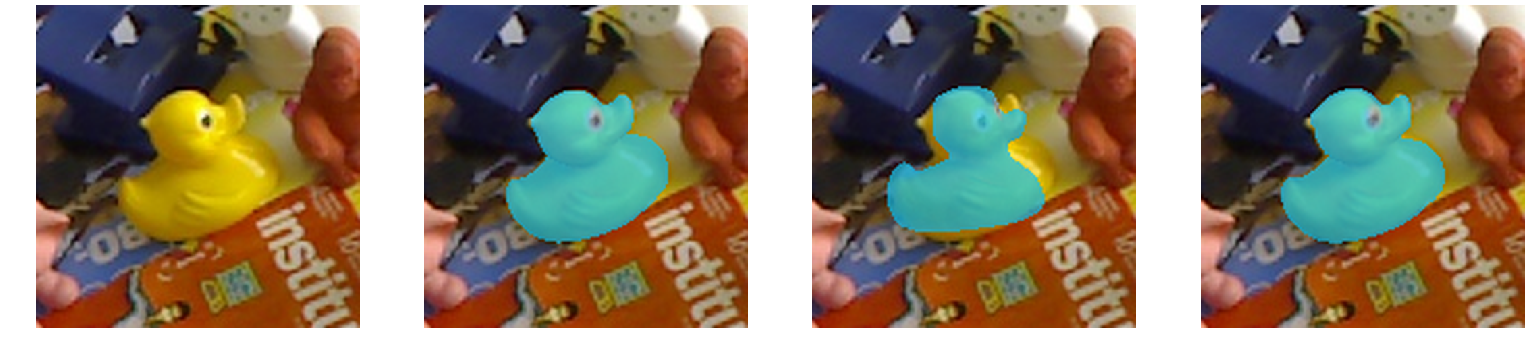}}\\
		\bmvaHangBox{\includegraphics[width=.45\linewidth]{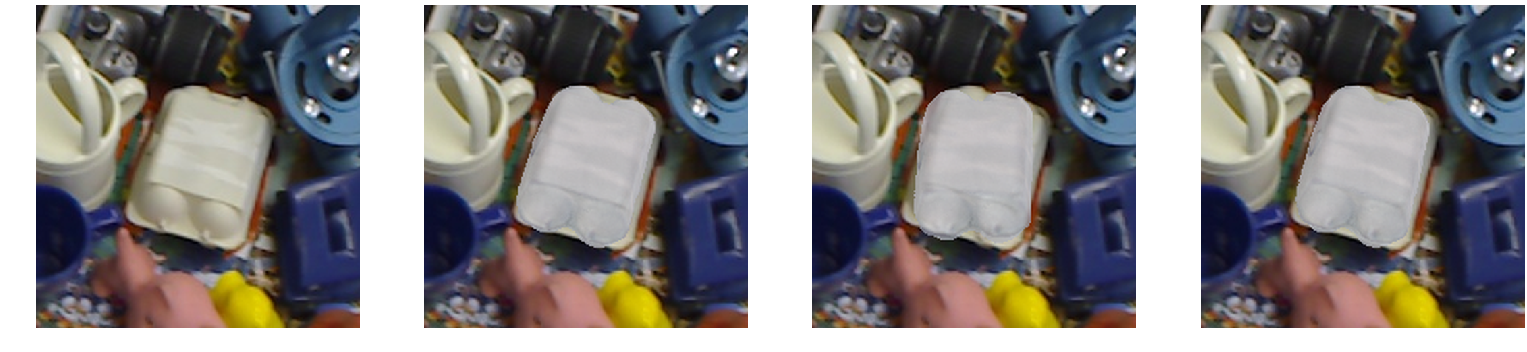}}&
		\bmvaHangBox{\includegraphics[width=.45\linewidth]{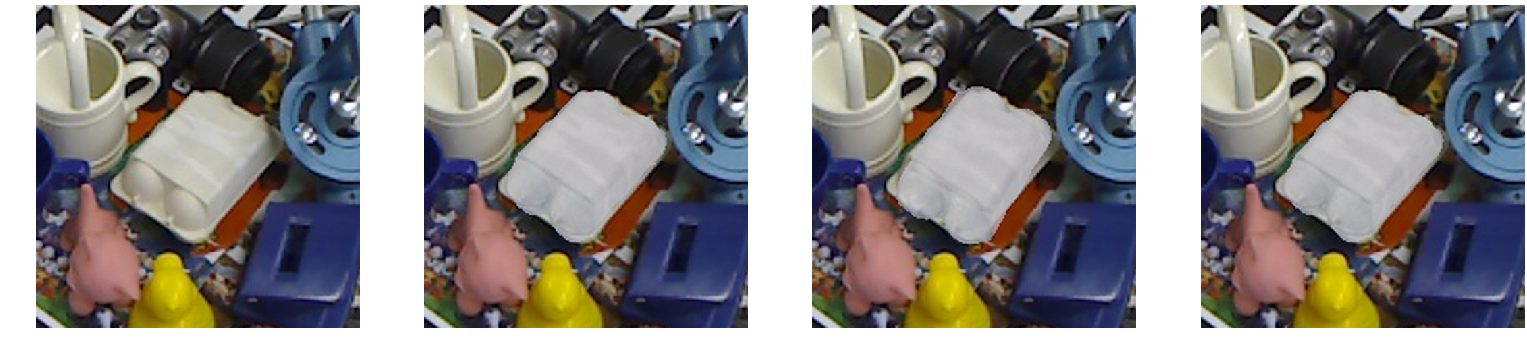}}\\
		\bmvaHangBox{\includegraphics[width=.45\linewidth]{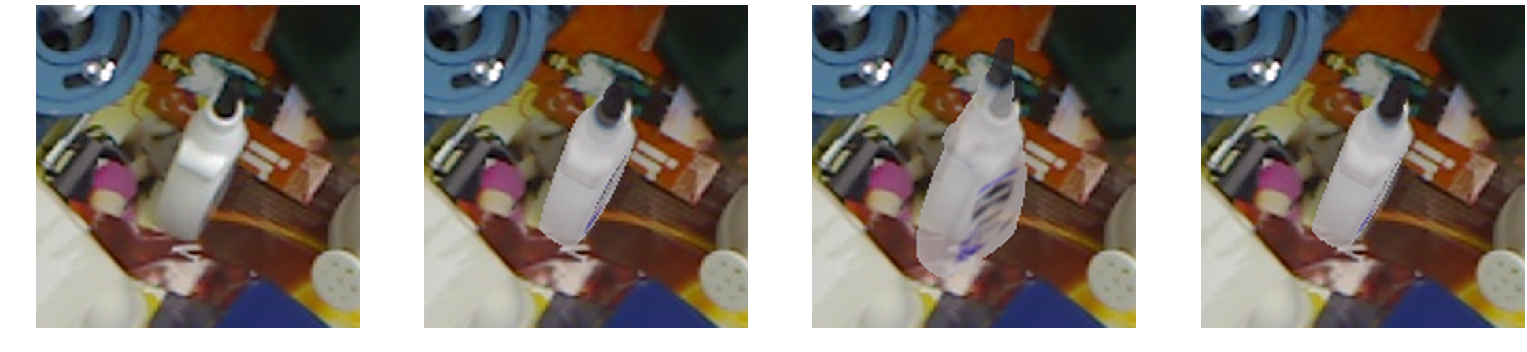}}&
		\bmvaHangBox{\includegraphics[width=.45\linewidth]{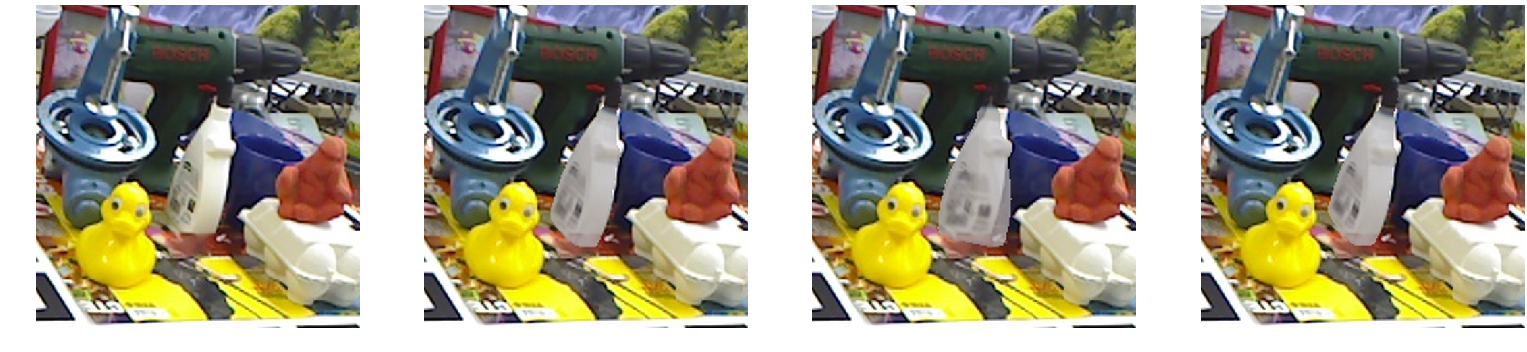}}\\
		\bmvaHangBox{\includegraphics[width=.45\linewidth]{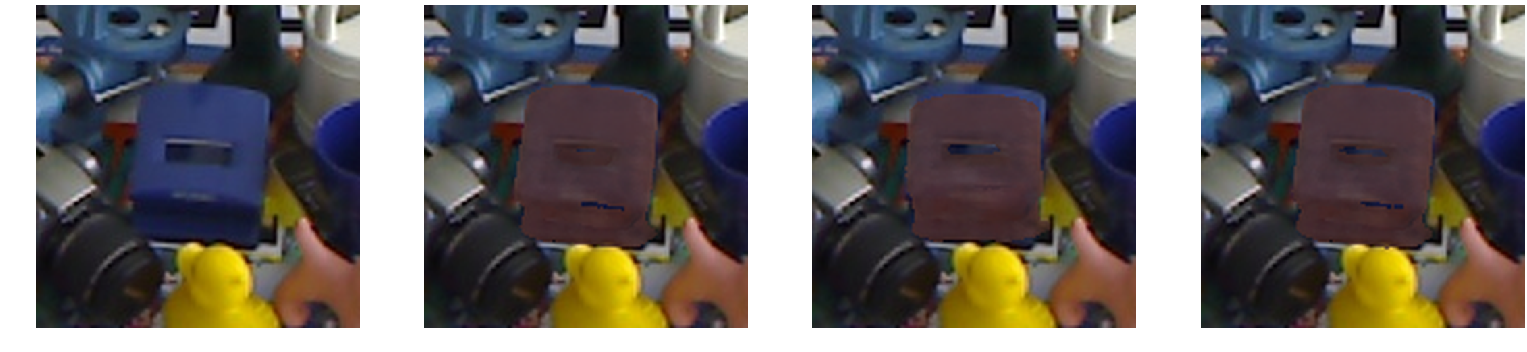}}&
		\bmvaHangBox{\includegraphics[width=.45\linewidth]{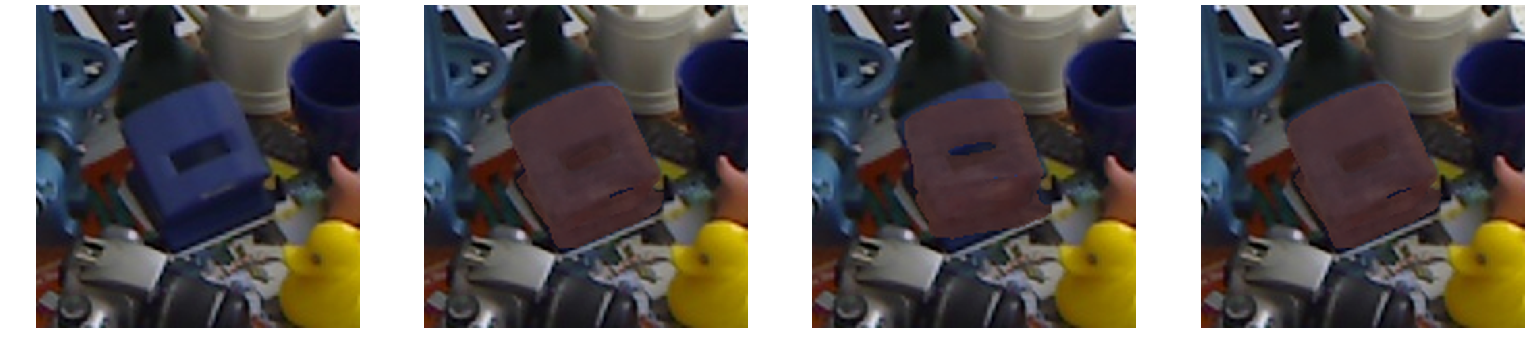}}\\
		\bmvaHangBox{\includegraphics[width=.45\linewidth]{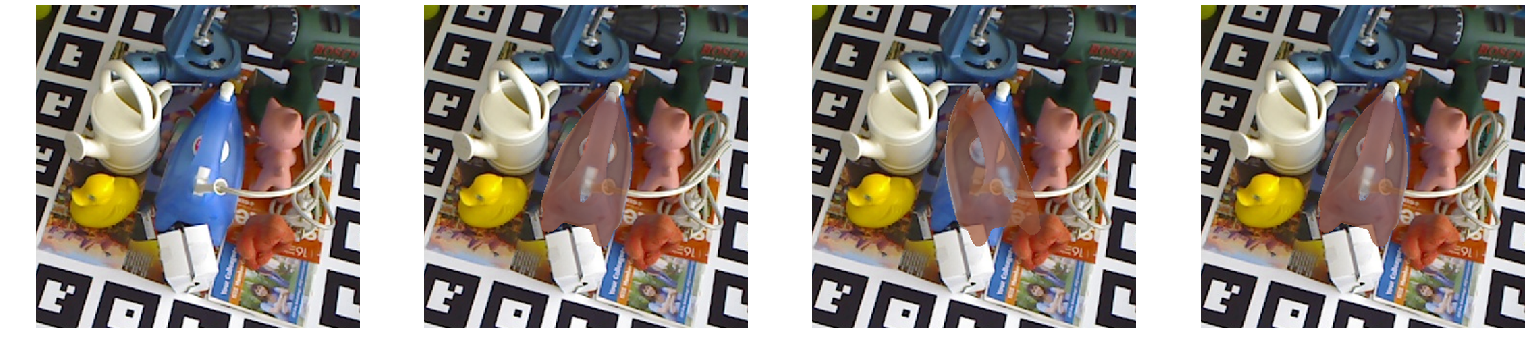}}&
		\bmvaHangBox{\includegraphics[width=.45\linewidth]{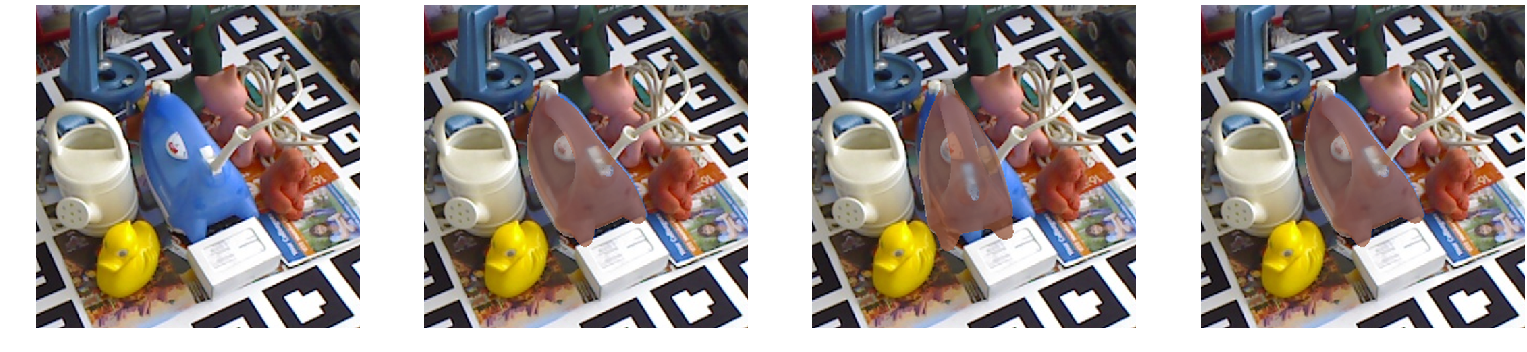}}\\
		\bmvaHangBox{\includegraphics[width=.45\linewidth]{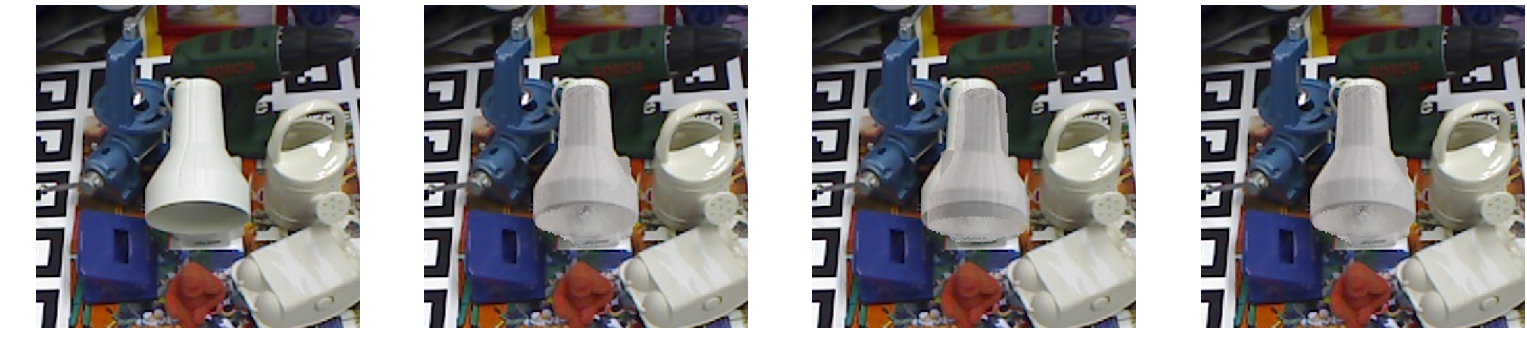}}&
		\bmvaHangBox{\includegraphics[width=.45\linewidth]{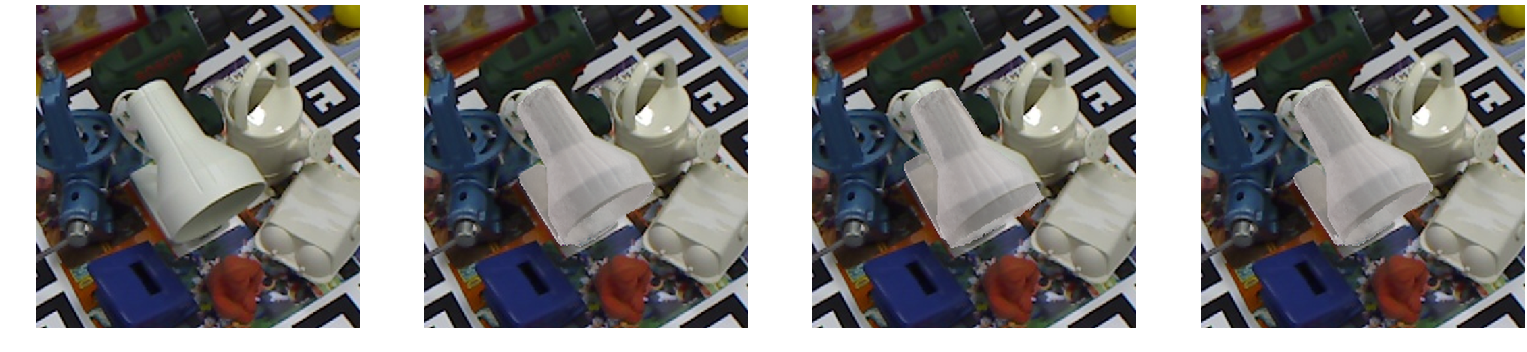}}\\
		\bmvaHangBox{\includegraphics[width=.45\linewidth]{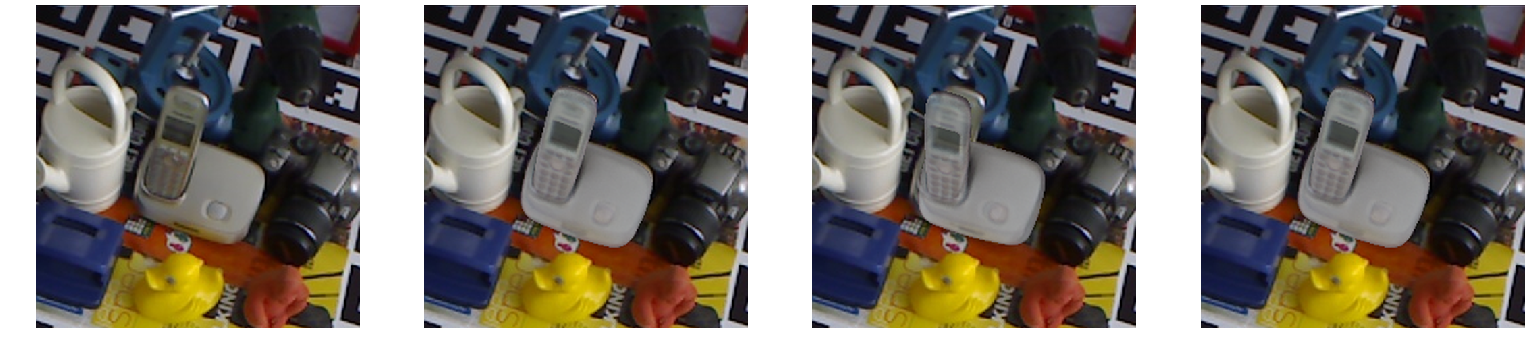}}&
		\bmvaHangBox{\includegraphics[width=.45\linewidth]{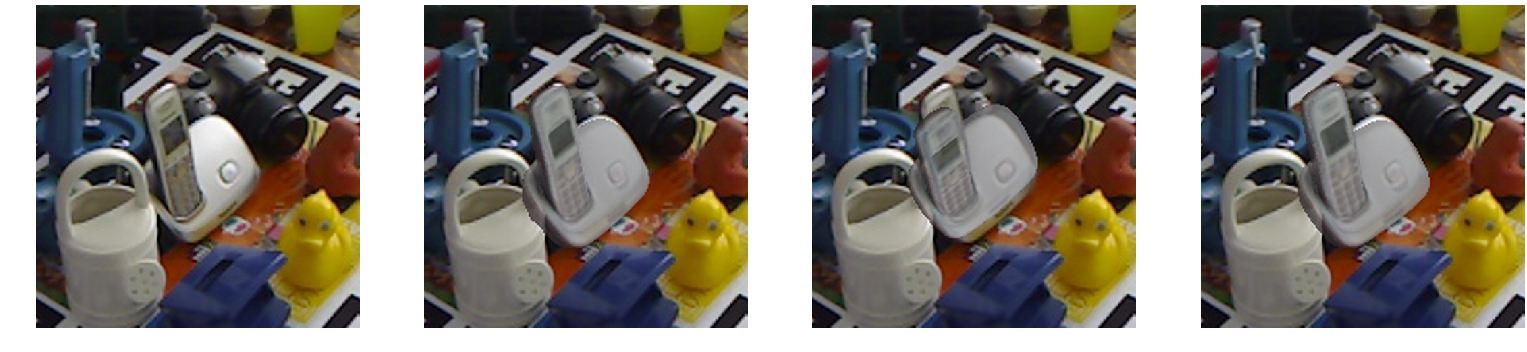}}\\
	\end{tabular}
	\vspace{1mm}
	\caption{Visual results of object pose estimation on LINEMOD~\cite{hinterstoisser2012LINEMOD}. For each sample, the four columns from left to right represent: the input image, the correct shape and orientation, our initial estimate and the final estimate after refining our initialization with DeepIM~\cite{li2018deepim}.}
	\vspace{-3mm}
	\label{fig:vis_linemod}
\end{figure}

Some qualitative results for 13 LINEMOD objects are shown in Figure \ref{fig:vis_linemod}. Given object image and its shape, our approach gives a coarse pose estimate which is then refined by pose refinement method given by DeepIM~\cite{li2018deepim}. 

\end{document}